%% file: paper.tex
\documentclass[twoside,11pt]{article}

\usepackage{jmlr2e}
\usepackage{graphicx}
\usepackage{subfigure}
\usepackage{amsmath}
\usepackage{amssymb}
\usepackage{cleveref}
\usepackage{tikz}
\usepackage{hyperref}
\usepackage{url}
\usepackage{pifont}
\usepackage{colortbl}
\usepackage{xcolor}
\usepackage{makecell}
\usepackage{fontawesome}
\usepackage{multirow}
\usepackage{wasysym}
\usepackage{makecell}
\usepackage{multicol}
\usepackage{paralist}
\usepackage{tcolorbox}
\usepackage{listings}
\usepackage{xcolor, soul}
\usepackage{adjustbox}

\makeatletter
\newcommand*{\rom}[1]{\expandafter\@slowromancap\romannumeral #1@}
\makeatother

\usepackage{lastpage}
\jmlrheading{26}{2025}{1-\pageref{LastPage}}{1/24}{1/25}{24-0043}{Pleines, Pallasch, Zimmer and Preuss}

\ShortHeadings{Endless Tasks to Benchmark Memory Capabilities of Agents}{Pleines, Pallasch, Zimmer and Preuss}
\firstpageno{1}

\begin{document}
\title{Memory Gym: Towards Endless Tasks to Benchmark Memory Capabilities of Agents}

\author{\name Marco Pleines \email marco.pleines@tu-dortmund.de \\
       \addr Department of Computer Science\\
       TU Dortmund University\\
       Dortmund, 44227, Germany
       \AND
       \name Matthias Pallasch \email matthias.pallasch@tu-dortmund.de \\
       \addr Department of Computer Science\\
       TU Dortmund University\\
       Dortmund, 44227, Germany
       \AND
       \name Frank Zimmer \email frank.zimmer@hochschule-rhein-waal.de \\
       \addr Department of Communication and Environment\\
       Rhine-Waal University of Applied Sciences\\
       Kamp-Lintfort, 47475, Germany
       \AND
       \name Mike Preuss \email m.preuss@liacs.leidenuniv.nl \\
       \addr Leiden Institute of Advanced Computer Science\\
       Universiteit Leiden\\
       EZ Leiden, 2311, The Netherlands}

\editor{Martha White}

\maketitle

\input{content/abstract}
\newpage
\input{content/intro}
\input{content/related_work}
\input{content/environments}
\input{content/baselines}

\input{content/experiments}
\input{content/conclusion}

\section*{Acknowledgements}
Our sincere appreciation goes to Vincent-Pierre Berges and Fabian Ostermann for their enlightening discussions that greatly enriched our work.
We also extend our gratitude to Andrew Lampinen for his invaluable insights on utilizing Transformer-XL as episodic memory.
Special thanks are due to Günter Rudolph for his unwavering support.
This project would not have been possible without the generous computing time provided by the Paderborn Center for Parallel Computing (PC2) and the support from the Linux-HPC-Cluster (LiDO3) at TU Dortmund.
We are deeply thankful for all of their contributions.

\newpage
\appendix
\input{content/annex}


\newpage

\vskip 0.2in
\bibliography{bibliography.bib}

\end{document}

%% file: content/abstract.tex

\begin{abstract}%
Memory Gym presents a suite of 2D partially observable environments, namely Mortar Mayhem, Mystery Path, and Searing Spotlights, designed to benchmark memory capabilities in decision-making agents.
These environments, originally with finite tasks, are expanded into innovative, endless formats, mirroring the escalating challenges of cumulative memory games such as ``I packed my bag''.
This progression in task design shifts the focus from merely assessing sample efficiency to also probing the levels of memory effectiveness in dynamic, prolonged scenarios.
To address the gap in available memory-based Deep Reinforcement Learning baselines, we introduce an implementation within the open-source CleanRL library that integrates Transformer-XL (TrXL) with Proximal Policy Optimization.
This approach utilizes TrXL as a form of episodic memory, employing a sliding window technique. 
Our comparative study between the Gated Recurrent Unit (GRU) and TrXL reveals varied performances across our finite and endless tasks.
TrXL, on the finite environments, demonstrates superior effectiveness over GRU, but only when utilizing an auxiliary loss to reconstruct observations.
Notably, GRU makes a remarkable resurgence in all endless tasks, consistently outperforming TrXL by significant margins.\\
Website and Source Code: \url{https://marcometer.github.io/jmlr_2024.github.io/}
\end{abstract}


\begin{keywords}
    deep reinforcement learning, actor-critic, memory, transformer, recurrence
\end{keywords}

%% file: content/intro.tex
\section{Introduction}

Imagine embarking on a long-awaited family vacation, with the open road stretching ahead.
As the car hums along, a lively atmosphere envelops the passengers, young and old alike, as they engage in a playful game of ``I packed my bag''.
Each family member takes turns adding an item to an imaginary bag, attempting to remember and recite the growing list correctly.
However, as the game unfolds and the list lengthens, the passengers encounter mounting challenges illustrating the finite nature of their individual memories.
Despite best efforts, varying strategies, and different capabilities, the weight of the growing list eventually leads to resignation.

Memory is not just a game – it is a critical tool for intelligent decision-making under imperfect information and uncertainty.
Without the ability to recall past experiences, reasoning, creativity, planning, and learning may become elusive.
In the realm of autonomously learning decision-making agents as Deep Reinforcement Learning (DRL), the agent's memory involves maintaining a representation of previous observations, a knowledge bank that grounds its next decision.
Memory mechanisms, be it through recurrent neural networks \citep{Rumelhart1986} or transformers \citep{Vaswani2017attention}, have enabled these agents to master tasks both virtual and real.
For instance, DRL methods have conquered complex video games as Capture the flag \citep{Jaderberg2018}, StarCraft \rom{2} \citep{Vinyals2019Starcraft2}, and DotA 2 \citep{Berner2019Dota2}.
Their success extends beyond virtual environments to real-world challenges as dexterous in-hand manipulation \citep{Andrychowicz2020dexterity} and controlling tokamak plasmas \citep{Degrave2022tokamak}.
These non-Markovian examples underscore that effective memory mechanisms are crucial; without them, the tasks are fundamentally unsolvable as they require the robust maintenance and manipulation of information over time.

\subsection{Endless Tasks Benchmark Memory Effectiveness and not just Efficiency}

To discover the most effective memory approach for complex tasks, as mentioned earlier, insightful benchmarks are needed.
However, conventional DRL memory benchmarks, which typically revolve around finite tasks that terminate upon reaching a terminal state regardless of success or failure, fall short in providing the necessary depth.
Finite tasks impose a predetermined limit on the demands placed on an agent's memory.
When various memory strategies successfully complete a fixed task, their performance is typically assessed based on efficiency rather than effectiveness.
Here, ``efficiency'' refers to the number of agent-environment interactions required to achieve effective behaviors, commonly known as sample efficiency.
Although these methods may reach the same end goal, their true effectiveness and memory capabilities often remain uncertain.
For example, a memory mechanism that retains more information or sustains it over extended periods could prove more effective, potentially leading to higher scores, increased returns, better survival times, or improved win rates.
This distinction could identify it as a more robust and effective approach compared to others assessed.

Cumulative memory games, such as ``I Packed My Bag'' and ``Simon''\citep{MorrisonBaer1977_Simon}, overcome the limitations of finite tasks by introducing an automatic curriculum through their inherently progressive and unbounded nature.
These games represent a special instance of endlessness: while theoretically offering a truly endless challenge, in practice, episodes inevitably terminate due to the finite nature of memory.
The game's expansion relies on the agents' continuous success, thereby embodying the concept of ``endlessness'' in tasks.
This approach provides deeper insights into the effectiveness of decision-makers when facing escalating demands for memory retention and recall.
Consequently, these challenges rigorously test an agent's memory, evaluating both the quantity of information it can retain and the duration for which it can robustly maintain it.

\subsection{Contributions: Novel Memory Benchmark and Transformer-XL Baseline}

To exploit the endless behavior of cumulative memory games to thoroughly benchmark memory effectiveness, we enhance our prior work Memory Gym \citep{pleines2023memory}, an open-source benchmark, designed to challenge memory-based DRL agents to memorize events across long sequences, generalize, be robust to noise, and be sample efficient.
Memory Gym encompasses three distinct finite environments: Mortar Mayhem, Mystery Path, and Searing Spotlights.
Each environment presents visual observations, multi-discrete action spaces, and crucially, they cannot be mastered without memory, demanding frequent memory interactions.
Building on this foundation, we advance Memory Gym by evolving each environment into an endless format, inspired by the ``I packed my bag'' game, to further test and develop memory capabilities in DRL agents.
To our understanding, no existing DRL memory benchmark possesses such endless tasks.

We further contribute an easy-to-follow baseline implementation of Transformer-XL (TrXL) \citep{Dai2019TrXL} to the open-source CleanRL library \citep{cleanRL2022}, based on the widely used Deep Reinforcement Learning (DRL) algorithm Proximal Policy Optimization (PPO) \citep{Schulman2017}.
A recurring theme in the realm of memory-based DRL is the lack of accessible and reproducible baselines.
Many significant works, particularly those utilizing transformers \citep{fortunato2019generalization, parisotto2020, Hill2021recon, Lampinen2021htmBallet, Parisotto2021ALD}, have not made their implementations publicly available.
As implementation details of DRL algorithms are vital \citep{Engstrom2020,andrychowicz2021what,huang2022ppo}, this lack of transparency poses significant challenges for the research community, hindering progress and reproducibility.
Therefore, our accessible implementation can serve as a beacon for the community fostering further research.

In our experimental analysis, we benchmark a TrXL and a recurrent agent, based on the Gated Recurrent Unit (GRU) \citep{Cho2014GRU}, across the original (finite) and the new endless environments of Memory Gym.
Our findings indicate that TrXL, only with the help of observation reconstruction as auxiliary loss, is most effective on the finite environments.
To our greatest surprise, in the endless environments, GRU consistently surpasses TrXL by large margins.

\subsection{Overview}

This paper begins with related work, followed by illustrating the dynamics in Memory Gym's environments, particularly focusing on their endless formats.
We detail the actor-critic model architecture, including the loss functions and key elements of the TrXL baseline. 
The subsequent section presents our experimental analysis, showcasing results from both finite and endless tasks.
We further explore TrXL's underperformance in endless tasks, providing insights and suggestions for future research.
Before concluding, we falsify our initial hypothesis about recurrence's vulnerability to spotlight perturbations in Searing Spotlights, emphasizing the significant impact of not normalizing estimated advantages during agent optimization.

%% file: content/related_work.tex
\section{Related Work}

Previous studies have explored the use of memory-based agents in a variety of tasks.
Some of them are originally fully observable but are turned into partially observable Markov Decision Processes by adding noise or masking out information from the agent's observation space \citep{Hausknecht2015DRQN, Heess2015, Meng2021, Shang2021}.
In the next subsection, we give a coarse overview of recently used benchmarks chronologically.
Subsequently, we discuss how these benchmarks primarily assess efficiency, highlighting a limitation in their scope, which we address with our contribution of Memory Gym's endless environments.

\subsection{An Overview of Memory Benchmarks in Deep Reinforcement Learning}

\textbf{Deepmind Lab 30} \citep{DMLab30_2016} presents a collection of 30 procedurally generated first-person 3D environments.
Starting from a general motor-control navigation task and visual observations, each environment poses a different goal such as collecting fruit, playing laser tag or traversing a maze.
\cite{pasukonis2023memmaze} have identified that agents can exploit the environments' skyboxes, which reduces the demands on memory.
\\\\
\noindent
\textbf{Minigrid} \citep{Chevalier-Boisvert2018minigrid} includes a 2D grid memory task inspired by the T-Maze \citep{Wiestra2007}.
In this environment, the agent is required to memorize an initial goal cue, traverse a long alley, and correctly choose the exit once the end is reached.
\\\\
\noindent
\textbf{Miniworld} \citep{gym_miniworld} encompasses a set of first-person 3D environments sharing similarities to the tasks introduced by Minigrid.
\\\\
\noindent
\textbf{VizDoom} \citep{wydmuch2018vizdoom} features first-person shooter 3D environments that revolve around motor-control navigation tasks, with the added challenge of computer-controlled enemies that may confront the agent.
\\\\
\noindent
The \textbf{Memory Task Suite} \citep{fortunato2019generalization} includes diverse memory-based environments across four categories: PsychLab \citep{Leibo2018psychlab} for image-based tasks (e.g. detect change), Spot the Difference for cue memorization, goal navigation tasks inspired by the Morris water maze \citep{DHooge2001watermaze}, and transitive object ordering tasks.
\\\\
\noindent
\textbf{Procgen} \citep{ProcgenCobbe202} consists of 16 2D environments that offer procedurally generated and fully observable levels.
These environments encompass a combination of puzzle-solving, platforming, and action-based games.
6 of these environments have been modified to become partially observable by reducing and centering the agent's visual observation.
\\\\
\noindent
\textbf{Numpad} \citep{parisotto2020} requires the agent to accurately press a series of keys in a specific sequence of fixed length.
The sequence is not provided to the agent, resulting in the agent using a trial-and-error approach while memorizing the underlying sequence.
\\\\
\noindent
\textbf{Memory Maze (1)} \citep{parisotto2020} shares similarities with the Morris Water maze \citep{DHooge2001watermaze}, as the agent must find an apple, then undergoes random repositioning, requiring it to relocate the apple's location.
\\\\
\noindent
\textbf{Dancing the Ballet} (Ballet) \citep{Lampinen2021htmBallet} presents a sequence of up to 8 dances visually, with the agent remaining stationary.
After all the dances are displayed, the agent is required to identify a specific one to successfully complete the task.
\\\\
\noindent
\textbf{PopGym} \citep{morad2023popgym}, published at ICLR 2023 alongside Memory Gym \citep{pleines2023memory}, is a benchmark with 15 environments classified as diagnostic, control, noisy, game, or navigation. These are designed for rapid convergence, offering vector instead of visual observations.
\\\\
\noindent
\textbf{Memory Maze (2)} \citep{pasukonis2023memmaze} showcases procedurally generated 3D mazes, where agents, from a first-person perspective, are tasked with locating specific objects as requested by the environment. Additionally, the authors have collected a dataset tailored for offline reinforcement learning.

\subsection{Conventional Benchmarks are Limited to Measuring Efficiency}
\label{sec:efficiency}

All of the aforementioned environments conclude with terminal states, regardless of whether the agent succeeds or fails.
Navigational tasks such as those found in Deepmind Lab 30, Minigrid, Miniworld, VizDoom, Procgen, Numpad, Memory Maze (1), and Ballet end by the agent discovering an exit.
This closure also applies to the navigation challenges within the Memory Task Suite and PopGym.
Additional tasks in PopGym conclude after the agent plays through an entire deck of cards or reaches a distinct goal state in games such as Concentration, Battleship, and Minesweeper.
Memory Maze (2) and tasks in Psychlab terminate after a predefined number of steps.

Due to these clearly finite episodes, the demands on the agent's memory have an upper bound.
Once the agent solves a given task, their performance is primarily measured by their efficiency.
Efficiency here refers to the number of steps an agent requires to complete an episode.
In scenarios such as escaping a maze, a more efficient agent will use fewer steps, whereas even a random agent might eventually find the exit if given sufficient time.
In both the literature and our results on the finite environments of Memory Gym (Section \ref{sec:finite_results}), agents are typically compared based on the number of samples (i.e., agent-environment interactions) they use during training to achieve their performance.
But what if an agent's memory is stronger at recalling larger amounts of relevant information over extended horizons?
Efficiency metrics may fail to fully reveal the true memory capabilities of agents, including their capacity to retain larger amounts of information or sustain that information over extended periods.

While the related works still offer potential for scaling in terms of memory demands, we take inspiration from cumulative memory games to expand Memory Gym to provide endless tasks that automatically and boundlessly test memory capabilities.
Identifying particularly strong memory mechanisms may enable the tackling of complex tasks, as noted in the introduction.

    
    


    
    




%% file: content/environments.tex
\section{Endless Memory Tasks Inspired by ``I Packed My Bag''}

Memory Gym's environments offer a unique feature that sets them apart from  the related benchmarks: an endless design tailored to rigorously test memory capabilities.
This section first explores how exploiting the concept of cumulative memory games can serve as a valuable benchmark to assess memory effectiveness, not merely efficiency.
We then describe how we expand Memory Gym's finite environments into endless ones.

It is important to clarify that `endless' should not be confused with `open-ended'.
Memory Gym's endless environments are not open-ended in nature; they possess clear objectives and defined structures.
In contrast, open-ended environments offer agents freedom to develop unique skills and strategies without stringent constraints as demonstrated by the Paired Open-Ended Trailblazer algorithm \citep{wang2020poet} or platforms like Minecraft \citep{fan2022minedojo}.

\subsection{Evaluating Effectiveness through Cumulative Memory Games}
\label{sec:effectiveness_cmg}

As argued in Section \ref{sec:efficiency}, measuring sample efficiency may obscure the true effectiveness of an agent's memory capabilities, especially under the assumption that the evaluated baselines are sufficient to solve the given finite tasks.
If endless tasks are not considered, what alternative methods can be used to assess an agent's memory effectiveness?
There are two primary approaches: scaling the difficulty of the task or scaling the agent.

One approach involves progressively ablating the agent itself while maintaining a fixed task difficulty.
For instance, the dimensions of transformer architectures or recurrent cells are scalable in terms of their parameter count, which is indicative of the memory's capacity.
By iteratively reducing these dimensions, a threshold is eventually reached beyond which the agent can no longer perform effectively.
Thus, with a fixed task, we can only determine the minimal configuration that still allows the agent to be effective.

In contrast, incrementally increasing the task's difficulty can determine the point at which an agent fails, thereby exposing the agent's limits of effectiveness.
The field of Curriculum Learning (CL), a specialization of transfer learning, offers methods to automate this scaling process \citep{Narvekar2020_curricula}.
CL enhances training efficiency by sequentially transferring knowledge from simpler tasks to more challenging ones, ultimately addressing tasks that might be too complex to tackle from scratch.
While setting up curricula could introduce implementation overhead, such as hand-crafting tasks or scheduling transitions between tasks, cumulative memory games naturally function out-of-the-box as an automatic curriculum.
These endless games increase in difficulty as the agent progresses within an episode, using a trivial heuristic to adjust the challenge in an online fashion, thereby eliminating the need for a separate task scheduler.

Applying the concept of cumulative memory games to a DRL memory benchmark provides significant insights.
Agents that excel at these tasks demonstrate robustness in retaining and recalling vast amounts of information over extended periods.
Moreover, because these games offer unbounded difficulty, they remain appropriately challenging—neither too difficult nor too easy—and hence continue to provide the right challenge for evaluating future memory mechanisms.
Mastery of these memory-intensive games highlights an agent's suitability for real-world applications, where maintaining context and adapting to incremental information are essential.

\subsection{From Finite to Endless: Expanding Memory Gym's Environments}

Prior to the underlying work, we introduced Memory Gym, a set of finite tasks benchmarking DRL agents to strongly depend on frequent memory interactions \citep{pleines2023memory}.
Memory Gym consists of the environments Mortar Mayhem, Mystery Path, and Searing Spotlights.
All of them expose visual observations, capturing an area of $84 \times 84$ RGB pixels, to the agent and allow for multi-discrete actions.
Leveraging procedural content generation, each episode is restarted with a new level sampled from a distinct set of seeds.
Additionally, the environments' simulation speed measures multiple thousands of steps per second, which is faster than most related benchmarks (see Appendix \ref{sec:sim_speed}).
Originally, these tasks were designed with finite goals, which limited the scope of memory challenges presented to the agents.

In transforming these environments from finite to endless, we aim to enhance the challenge posed to an agent's memory by increasing the amount of information to memorize over expanding time:
\begin{itemize}
\item \textbf{Endless Mortar Mayhem} tasks the agent to memorize and execute an ever-expanding list of commands in the correct order.
\item \textbf{Endless Mystery Path} exhibits a never-ending, invisible path for the agent to traverse. Deviating from the path forces a restart from the beginning, requiring the agent to recall both the route and previous missteps accurately.
\item \textbf{Endless Searing Spotlights} demands the agent to track its obscured location while avoiding threatening spotlights and collecting coins.
\end{itemize}

For detailed configuration and scaling of these environments, refer to Annex \ref{sec:environment_parameters}, which presents the reset parameters.
The next subsections explore the transitions from finite tasks to endless ones.

\subsubsection{Endless Mortar Mayhem}

\begin{figure}
    \centering
    \subfigure[Agent Observation]{
        \includegraphics[width=0.275\textwidth]{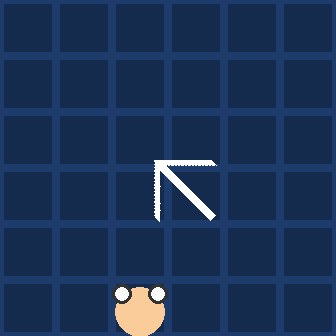}
        \label{fig:emm_a}
    }
    \hspace{1.0cm} 
    \subfigure[Command Observation and Execution]{
        \includegraphics[width=0.45\textwidth]{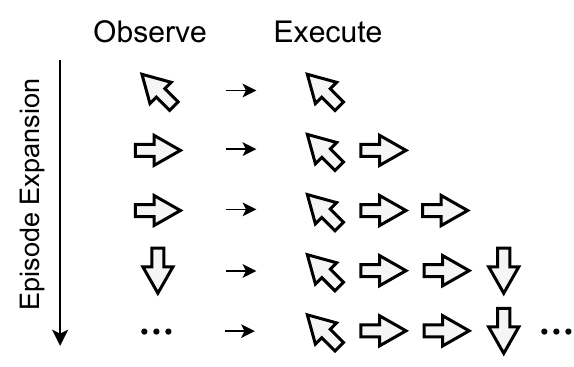}
        \label{fig:emm_b}
    }
    \caption{Endless Mortar Mayhem~(a) is observed from a top-down view and its arena occupies the entire screen. If the agent, depicted as a circle with white paws, walks off, it re-enters from the opposite side. Episodes  endlessly expand by alternately visualizing and executing commands (b), behaving as an automatic curriculum.
    }
    \label{fig:emm}
\end{figure}

The finite variant of Mortar Mayhem takes place inside a grid-like arena and comprises two tasks.
Initially, the agent is immobile and must memorize a sequence of ten commands, followed by executing each command in the observed order.
A command instructs the agent to move to an adjacent floor tile or remain at the current one.
Failure to execute a command results in episode termination, whereas each successful execution yields a reward of $+0.1$.

Endless Mortar Mayhem~(Figure \ref{fig:emm}) extends the to-be-executed command sequence continuously, enabling potentially infinite sequences.
To accommodate this, the command visualization and execution are alternated.
As shown in Figure \ref{fig:emm_b}, an episode begins by displaying one command, followed by its execution.
Subsequently, the next visualization reveals only the second command, requiring the agent to execute both the first and second commands.
This automatic curriculum allows the to-be-executed command sequence to continually expand, while each new command is presented only once.

\subsubsection{Endless Mystery Path}

\begin{figure}
    \centering
    \subfigure[Agent Observation]{
        \includegraphics[width=0.275\textwidth]{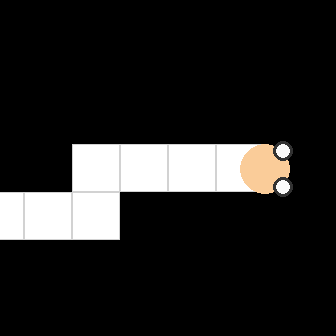}
        \label{fig:emp_a}
    }
    \subfigure[Endless Path]{
        \includegraphics[width=0.66\textwidth]{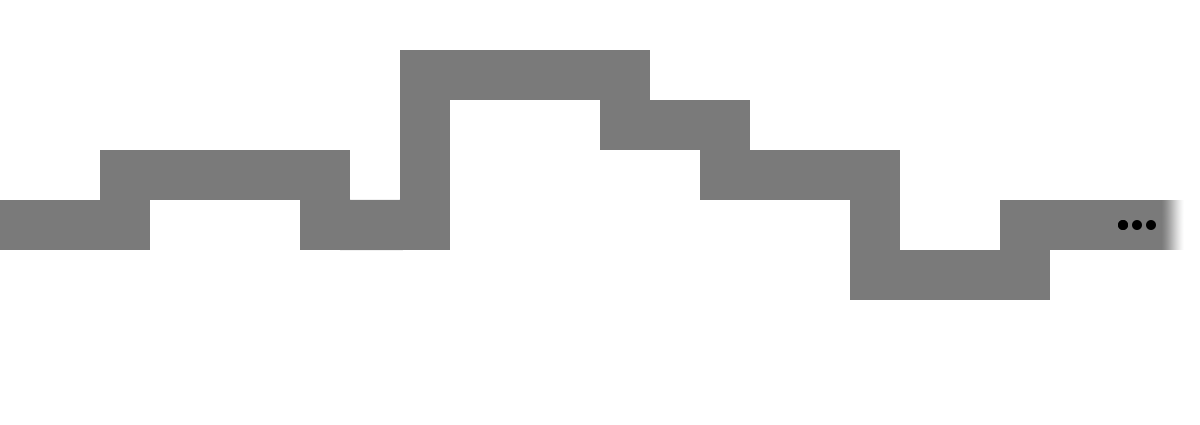}
        \label{fig:emp_b}
    }
    \caption{In Endless Mystery Path~(a), the agent observes only a segment of the environment at a time. By rendering the path tiles past it, the agent acquires a sense of horizontal movement, even though its horizontal position remains visually constant. Figure (b) showcases an example of a procedurally generated endless path.
    }
    \label{fig:emp}
\end{figure}

The core concept of the original Mystery Path environment requires the agent to navigate an invisible path until reaching its end.
If the agent strays off the path, it is relocated to the path's origin.
To succeed, the agent must consistently memorize and recall its steps along the path, as well as the locations where it fell off.

Endless Mystery Path (Figure \ref{fig:emp}) introduces a never-ending path, which is always generated from left to right.
Consequently, we eliminate the agent's ability to move left.
As the endless path cannot be fully captured in an $84 \times 84$ RGB pixel observation, we visually fix the agent's horizontal position.
If nothing but the agent is shown in the observation, the agent lacks information about its horizontal motion.
To address this, we render the path tiles behind the agent, providing visual cues of the local horizontal position.

Regarding terminal conditions, we aim to shorten episodes if the agent performs poorly.
Firstly, the agent is given a time budget of $20$ steps to reach the next tile.
This budget is reset when the agent moves to the next tile or falls off.
Another condition terminates the episode if the agent falls off before reaching its best progress.
Lastly, the episode is terminated if the agent falls off at the same location for a second time.
With these conditions in place, an episode can potentially last indefinitely if the agent consistently succeeds.

In terms of rewards, the agent earns $+0.1$ for each previously unvisited tile of the path it reaches.
While this may initially appear as a dense reward function, any misstep that causes the agent to fall off means it cannot reclaim those rewards.
The agent must retrace its steps and regain its prior progress to continue accumulating rewards.
The longer it takes to catch up, the sparser the rewards become.

\subsubsection{Endless Searing Spotlights}

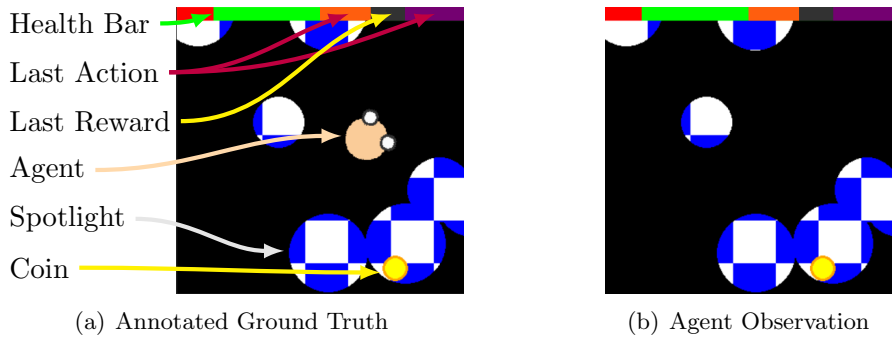
\begin{figure}
\begin{center}
\subfigure[Annotated Ground Truth]{\label{fig:ss_gt}
\begin{tikzpicture}
\definecolor{tw}{rgb} {0.9,0.9,0.9}
\definecolor{skin}{rgb} {1.0,0.85,0.67}
\node [anchor=west] (health) at (-0.35,3.6) {Health Bar};
\node [anchor=west] (action) at (-0.35,2.95) {Last Action};
\node [anchor=west] (reward) at (-0.35,2.3) {Last Reward};
\node [anchor=west] (agent) at (-0.35,1.65) {Agent};
\node [anchor=west] (spot) at (-0.35,1.0) {Spotlight};
\node [anchor=west] (coin) at (-0.35,0.35) {Coin};
\begin{scope}[xshift=2cm]
    \node[anchor=south west,inner sep=0] (image) at (0,0) {\includegraphics[width=0.25\textwidth]{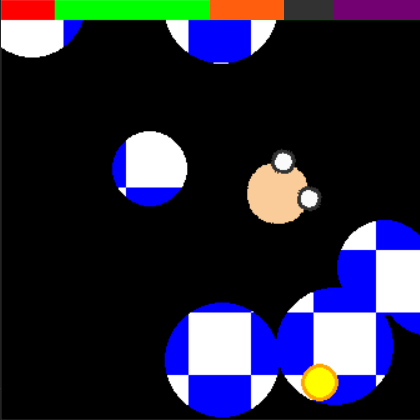}};
    \begin{scope}[x={(image.south east)},y={(image.north west)}]
        \draw [-latex, ultra thick, green] (health) to[out=0, in=200] (0.125,0.975);
        \draw [-latex, ultra thick, purple] (action) to[out=0, in=200] (0.6,0.975);
        \draw [-latex, ultra thick, purple] (action) to[out=0, in=200] (0.9,0.975);
        \draw [-latex, ultra thick, yellow] (reward) to[out=0, in=200] (0.75,0.975);
        \draw [-latex, ultra thick, skin]  (agent) to[out=0, in=180] (0.575,0.55);
        \draw [-latex, ultra thick, tw]  (spot) to[out=0, in=180] (0.375,0.15);
        \draw [-latex, ultra thick, yellow]  (coin) to[out=0, in=180] (0.71,0.075);
    \end{scope}
\end{scope}
\end{tikzpicture}
}
\hspace*{15mm}
\subfigure[Agent Observation]{\label{fig:ss_obs}\includegraphics[width=0.25\textwidth]{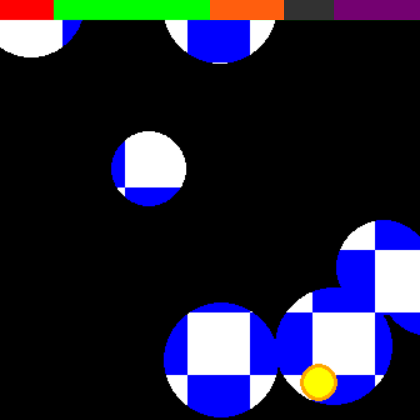}}
\caption{In Endless Searing Spotlights' annotated ground truth, the top rows of pixels display the agent's remaining health points, its last action, and indicate whether a positive reward was received during the last step.
The last action is encoded using two chunks and three colors, while the last positive reward is represented by two colors.
In the agent's observation~(b), the spotlights play a role in revealing or hiding other entities.}
\label{fig:ss}
\end{center}
\end{figure}

Endless Searing Spotlights (Figure \ref{fig:ss}) features a pitch-black environment illuminated only by moving and threatening spotlights.
The agent starts with ten health points, losing one each time it is hit by a spotlight.
At the beginning of each episode, the environment is fully lit, but it quickly dims to darkness within a few steps.
The episode ends if the agent's health points are depleted.
To survive, the agent must cleverly use the darkness as cover, relying on memory to remember past actions and previous positions to infer its current location.
The agent's observation space includes its last action, current health, and any positive reward received in the previous step.

To mitigate monotonous behaviors, a coin collection task is incorporated.
In the finite version, the agent aims to collect a randomly placed coin, worth $+0.25$, and then exit the level for an additional $+1$ reward. Effective memory use enables the agent to track its location, coin positions, and the exit.
In the endless concept, the exit is removed, and the coin collection mechanism is revamped to spawn a new coin each time one is collected.
This coin remains visible for six frames, while the agent has 160 steps to collect it.
Distinct from the finite version, spotlights now appear at a steady rate, rather than intensifying over time.

%% file: content/baselines.tex

\section{Transformer-XL Baseline}

In our experiments, we employ the widely-recognized Deep Reinforcement Learning (DRL) algorithm Proximal Policy Optimization (PPO) \citep{Schulman2017}.
To add memory abilities to PPO, either recurrent neural networks (RNN) or attention mechanisms (i.e. transformer) are potential solutions.
Prior works have extensively demonstrated the effectiveness of leveraging RNNs, such as Long Short-Term Memory (LSTM) \citep{Hochreiter1997LSTM} and Gated Recurrent Unit (GRU) \citep{Cho2014GRU}, within the realm of DRL \citep{mnih2016a3c, Esepholt2018IMPALA, huang2022ppo}.
This also applies to previous studies employing transformers, as Gated Transformer-XL (GTrXL) \citep{parisotto2020} and Hierarchical Chunk Attention Memory (HCAM) \citep{Lampinen2021htmBallet}.
However, integrating these memory-enhancing techniques into DRL algorithms is not straightforward, particularly due to unavailable transformer-based DRL implementations.
To bridge this gap, we contribute an easy-to-follow baseline implementation of Transformer-XL (TrXL) \citep{Dai2019TrXL} in PPO.
Before delving into the transformer-based PPO baseline, we describe the actor-critic model architecture and the losses used during training.

\subsection{Actor-Critic Model Architecture}

\begin{figure}
    \centering
    \includegraphics[width=0.675\textwidth]{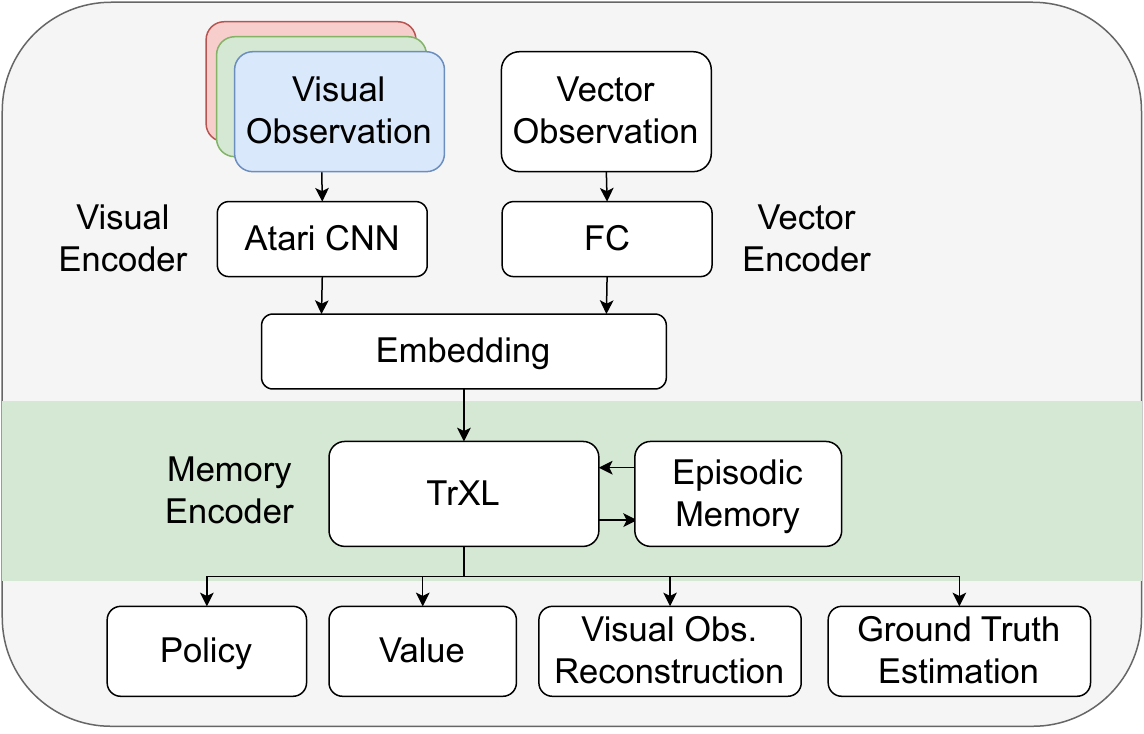}
    \caption{Overview of the actor-critic model architecture. The model features observation encoders and a memory encoder (green). The encoders' parameters are shared among several heads dedicated to the policy, state-value function, observation reconstruction, and ground truth estimation.}
    \label{fig:model}
\end{figure}

Figure \ref{fig:model} presents a broad overview of the actor-critic model architecture.
The encoding process begins with visual observations, and optionally, vector observations (i.e. game state information).
To encode visual observations (i.e. pixel observations) of shape $84\times84\times3$, we employ the Atari CNN \citep{mnih2015DQN}.
The vector observation is encoded by a fully connected layer (FC).
The outputs of these encoders are concatenated and subsequently embedded into the input space of the transformer-based memory encoder.
Once the data is propagated through the memory encoder, it is further processed by each individual head that may contain further fully connected hidden layers.
The policy head (actor) samples multi-discrete actions as required by Memory Gym, while the value head (critic) approximates the state-value.

Optionally, there is a head dedicated to reconstructing the original visual observation to strengthen the learning signal: the latent output of the memory encoder is fed to a transposed Atari CNN.
The utility of such an auxiliary head is demonstrated by some of our results (Section \ref{sec:experiments}) and prior works by \cite{Lampinen2021htmBallet} and \cite{Hill2021recon}.
During development, we also tried to attach the reconstruction head directly to the encoder, but this led to catastrophic results.
Another optional head establishes a diagnosis tool by estimating ground truth information, which is provided as labels by the environment.
Despite the injection of illegal information into the agent's training, we utilize this head in Section \ref{sec:weak_signal} to demonstrate that TrXL's performance improves in Endless Mortar Mayhem due to this richer learning signal, indicating that its overall low performance is not attributed solely to a lack of capacity.
Initially, the parameters of such a head can be optimized post-agent training to evaluate whether distinct information can be queried from the agent's memory \citep{Baker2020HideAndSeek}.

\subsection{Loss Function Composition}

Our baseline utilizes PPO's clipped surrogate objective \citep{Schulman2017}.
Due to leveraging memory, the to be selected action $a_t$ of the policy $\pi_\theta$ depends on the current observation $o_t$ and the memory encoder's output that we refer to as hidden state $h_t$.

\begin{equation}
    L^{C}_t(\theta) = -\mathbb{E}_t \left [ \text{min}\left(q_t(\theta)A_\pi^{\text{GAE}}(o_t, h_t, a_t),\,\text{clip}\left(q_t(\theta),1- \epsilon,1+\epsilon\right)A_\pi^{\text{GAE}}(o_t, h_t, a_t) \right) \right ]
    \label{eq:policy_loss}
\end{equation}
\begin{equation*}
    \textnormal{with ratio}~q_t(\theta) = \frac{\pi(a_t|o_t, h_t,\theta)}{\pi(a_t|o_t, h_t, \theta_{\text{old}})}
\end{equation*}
\noindent
$A_\pi^{\text{GAE}}$ denotes advantage estimates based on generalized advantage estimation (GAE) \citep{Schulman2016GAE}, $\theta$ the trainable parameters of a neural net, and $\epsilon$ the clip range.
$t$ depicts the current time step.
The value function is optimized using the clipped squared-error loss $L_t^V(\theta)$, which is apparent in the implementation of \cite{Schulman2017}:

\begin{equation}
    L^{VClip}_t(\theta) = \left(\text{clip}\left(V(o_t, h_t, \theta), V(o_t, h_t, \theta_{\text{old}})-\epsilon, V(o_t, h_t, \theta_{\text{old}}) + \epsilon\right) - \hat{V}_t\right)^2
\end{equation}

\begin{equation}
    L_t^{V}(\theta) = \max\left[\left(V(o_t, h_t, \theta) - \hat{V}_t\right)^2,~L^{VClip}_t(\theta)\right]
    \label{eq:value_clip}
\end{equation}
\noindent
The auxiliary observation reconstruction is optimized using the binary cross entropy:

\begin{equation}
    L^R_t(o_t, \hat{o}_t) = -[\hat{o}_t \log(o_t) + (1 - \hat{o}_t) \log(1 - o_t)],
\end{equation}
\noindent
and the optional ground truth estimation leverages the squared-error loss, which optimizes the concerned head's parameters $\phi$:

\begin{equation}
    L^Y_t(\hat{y_t},y_t,\phi) =  (\hat{y}_t - y_t)^2
\end{equation}
\noindent
The final combined loss is depicted by $L^{C+V+H+R+Y}_t(\theta)$:

\begin{equation}
	L^{C+V+H+R+Y}_t(\theta)=\mathbb{E}_t [L^{C}_t(\theta)+c_1L^{V}_t(\theta)-c_2\mathcal{H}[\pi_\theta](o_t, h_t) + c_3L^R_t(o_t, \hat{o}_t) + c_4 L^Y_t(\theta)]
\end{equation}
\noindent
where $\mathcal{H}[\pi_\theta](o_t,h_t)$ denotes an entropy bonus encouraging exploration \citep{Schulman2017}.
$c_1$ to $c_4$ are coefficients to scale each loss, excluding the clipped surrogate objective.

\subsection{Transformer-XL as Memory Mechanism}
\label{sec:trxl_baseline}

\begin{figure}[t!]
  \centering
  \subfigure[Episodic Memory]{
    \includegraphics[clip, trim=1.5cm 0 0.7cm 0,width=0.465\textwidth]{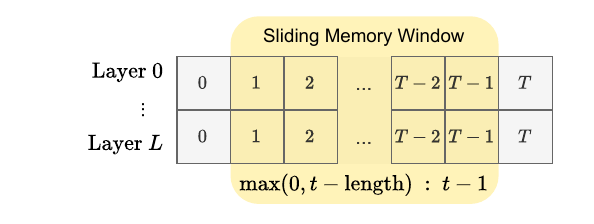}
    \label{fig:window}
  }
  \vspace{\baselineskip}
  \subfigure[Transformer-XL Layer]{
    \includegraphics[width=0.475\textwidth]{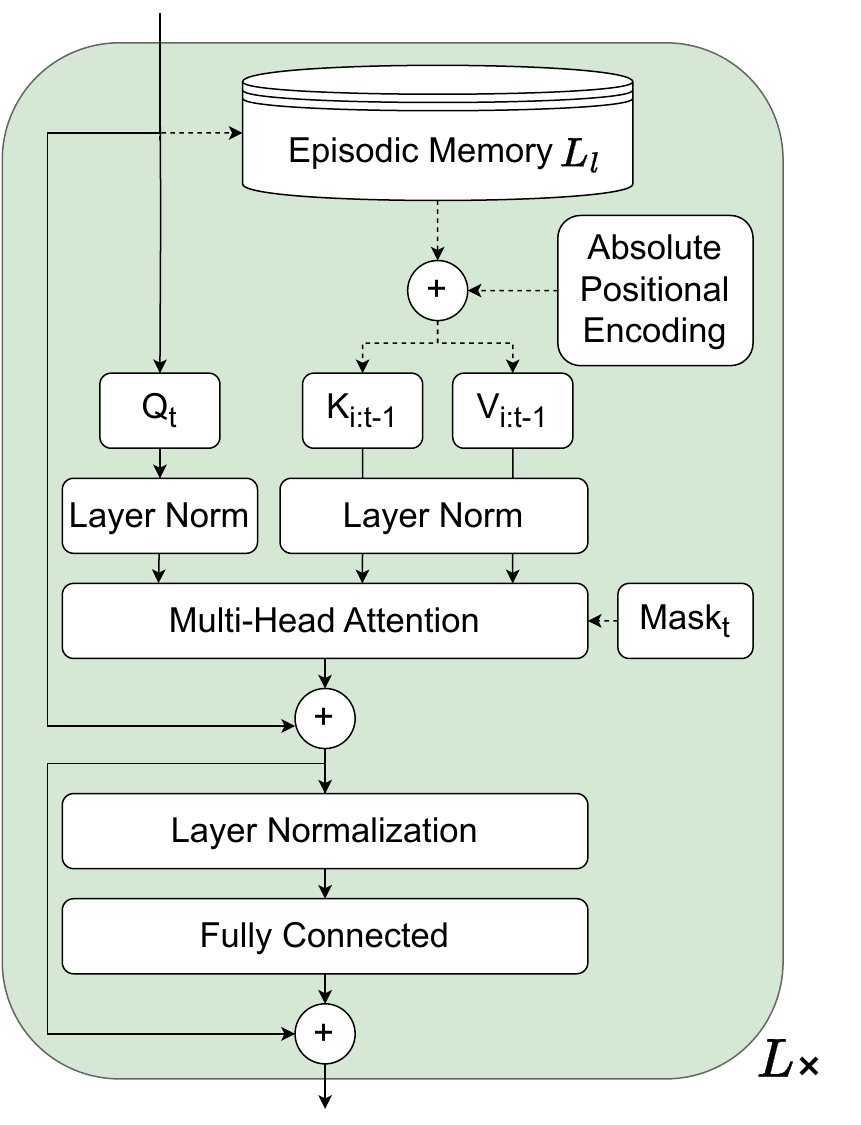}
    \label{fig:block}
  }
  \hfill
  \caption{Figure \subref{fig:window} illustrates the episodic memory that stores past inputs to all Transformer-XL layers for every step taken by the agent.
  Input sequences to the Transformer-XL layers are retrieved using a sliding memory window.
  $T$ denotes the episode length.
  Figure \subref{fig:block} depicts the architecture of the Transformer-XL block that stacks $L$ layers.
  The dashed lines state that there is no gradient flow into the episodic memory, the absolute positional encoding, and the mask.
  }
  \label{fig:trxl}
\end{figure}

When processing sequential data, transformers employ attention mechanisms to capture global dependencies in parallel across the entire sequence \citep{Vaswani2017attention}, while RNNs iteratively propagate information through recurrent connections.
Our transformer baseline draws conceptual inspiration from TrXL \citep{Dai2019TrXL}, GTrXL \citep{parisotto2020}, and HCAM \citep{Lampinen2021htmBallet}.
The original transformer architecture proposed by \cite{Vaswani2017attention} is designed as a sequence-to-sequence model with an encoder-decoder structure.
However, for the purpose of DRL, we adapt the architecture to a sequence-to-one model, focusing solely on the encoder as done in GTrXL and HCAM.
This modification is driven by the underlying Markov Decision Process, which maps the environment's current state to a single action rather than a sequence of actions.

To begin with, it needs to be clarified what kind of sequential data is fed to a transformer, which is typically composed of multiple stacked layers.
Naively, the input data can be conceptualized as a sequence of observations.
In the context of visual observations, this is commonly referred to as frame stacking, where consecutive frames are combined.
However, instead of expensively stacking raw frames, we leverage the past (i.e. cached) outputs of the model's embedding module (Figure \ref{fig:model}), which serves as the input sequence to the first transformer layer.
The subsequent layers make use of the past outputs from their preceding layers.
These sequences, those elements we refer to as cached hidden states, are stored in the agent's episodic memory (Figure \ref{fig:window}) and are collected during inference based on the episode's current timestep $t$.
To accommodate computational challenges of potentially endless episodes, we employ a sliding memory window approach, where the input sequence of these hidden states is derived from a fixed-length window.
By leveraging cached hidden states in the input sequence, the agent's memory facilitates the ability to attend to events that extend beyond the boundaries of the fixed-length memory window, aligning with the segment-level recurrence of TrXL \citep{Dai2019TrXL}.
Let \( N \) represent the number of layers and \( L \) denote the window length.
The maximum context length \( L_{\text{max}} \) is given by

\begin{equation}
    L_{\text{max}} = N \times (L - 1) + 1\text{.}
    \label{eq:length}
\end{equation}
Note that gradients are not back propagated through the episodic memory because of caching.
Once propagated through all transformer layers, a final hidden state is produced to inform the agent's policy.

The architecture of our TrXL encoder block is depicted in Figure \ref{fig:block}.
At each timestep $t$, the block receives an input that is solely based on the current timestep.
This input is added to the episodic memory and serves as the query $Q_t$ during the execution of multi-head attention (MHA) \citep{Vaswani2017attention}.
The keys $K$ and values $V$ used in MHA are derived from the same data based on self-attention.
Specifically, we retrieve $K$ and $V$ by slicing the episodic memory using the bounding indices $i = \max(0, t - \text{{window length}})$ and $t - 1$.
To ensure the positional information of $K$ and $V$ remains coherent, we add a positional encoding following the approach from \cite{Vaswani2017attention}.
We use the absolute variant as a default as we found a learned one to be effective, but less sample efficient.
The underlying positional encoding matrix is scaled to the maximum episode length and not to the length of the window.
Therefore, every sequence element of an episode is associate with its distinct position.
In the case of endless episodes, the length of the positional encoding is fixed to 2048.

To restrict attention to time steps up to the current timestep $t$, we employ a strictly lower triangular matrix as a mask in MHA.
This mask is only necessary when $t$ is smaller than the length of the sliding memory window.
The remaining configuration of the block adheres to the findings of \cite{parisotto2020}, which suggest improved performance with pre layer normalization and the identity map reordering.
We encountered vanishing gradient issues when applying layer normalization after MHA and the fully connected layer, which is referred to as post layer normalization.
Although our implementation supports the gating mechanism of GTrXL, it resulted in either lower or equivalent performance while being computationally more expensive (Appendix \ref{sec:gtrxl_lstm}).

%% file: content/experiments.tex
\section{Experimental Analysis}
\label{sec:experiments}

We run empirical experiments on Memory Gym using the just introduced TrXL baseline and one based on GRU, both powered by Proximal Policy Optimization \citep{Schulman2017}.
The next subsection explains why GRU and TrXL are the only baselines considered.
Next, we detail the evaluation protocol.
Section \ref{sec:finite_results} presents results on the finite tasks, showing that TrXL exhibits stronger policies than GRU, but only when leveraging observation reconstruction as an auxiliary loss.
Contradicting our initial expectations and the results on the finite environments, we unveil surprising findings that highlight GRU's superiority over TrXL in all endless environments.
Afterward, we compare the results from the finite and endless environments, demonstrating that the finite settings offer only a limited view of the agents' true memory capabilities.
We then broadly explore several hypotheses on why TrXL might be less effective.
Finally, we falsify that recurrent agents are vulnerable to spotlight perturbations when using enhanced hyperparameters (e.g., not normalizing advantage estimates).

\subsection{Considered Baselines}
\label{sec:considered_baselines}

Other related and significant baselines, such as the Memory Recall Agent (MRA) \citep{fortunato2019generalization}, Hierarchical Chunk Attention Mechanism (HCAM) \citep{Lampinen2021htmBallet}, and Gated Transformer-XL (GTrXL) \citep{parisotto2020}, are not considered due to their closed-source nature, which hampers reproduction.
Unlike our previous study \citep{pleines2023memory}, we exclude HELM (History comprEssion via Language Models) \citep{Paischer2022HELM} and its successor, HELMv2 \citep{paischer2022helmv2}.
This decision stems from HELM's suboptimal wall-time efficiency, which is six times more expensive than that of a GRU agent \citep{pleines2023memory}, making hyperparameter optimization impractical for our purposes.
Given our strict compute budget of 25,000 GPU hours, we dedicated our resources to developing the endless environments and the TrXL baseline instead.

For a comprehensive overview of the hyperparameters used in our experiments, see Appendix \ref{sec:hyperparameters}.
We also conduct preliminary tests using GTrXL and Long Short-Term Memory (LSTM) \citep{Hochreiter1997LSTM} on the finite environments without extensive hyperparameter tuning.
GTrXL is originally based on the DRL algorithm V-MPO \citep{Song2020_VMPO}.
We modify our TrXL baseline to adopt the gating mechanism.
Results for GTrXL and LSTM on the finite environments are provided in Appendix \ref{sec:gtrxl_lstm}.
These models usually underperformed when applied directly, suggesting that out-of-the-box applications might necessitate hyperparameter optimization.

Additionally, we reaffirm that Memory Gym's environments require memory by training naive baselines, such as frame stacking (Appendix \ref{sec:minor_baselines}).
A selection of wall-time efficiency statistics is available in Appendix \ref{sec:wall_time}.

\subsection{Evaluation Protocol}
\label{sec:evaluation_protocol}

A major challenge in DRL research is the statistical uncertainty resulting from high computational costs and the stochastic nature of training algorithms \citep{agarwal2021statistics, white2024}.
Due to these expenses, many studies rely on limited samples, such as those evaluating PPO, MRA, and HCAM with only three runs (i.e., seeds) per experiment.
While our empirical protocol does not fully resolve this issue, it mitigates it by repeating each experiment five times on independent seeds.
The code and data behind all plots are publicly available at \url{https://marcometer.github.io/jmlr_2024.github.io/}.

The forthcoming results on the finite and endless environments (Sections \ref{sec:finite_results} and \ref{sec:endless_results}) display the mean performance of each individual run.
Instead of aggregating these runs, we take advantage of the available space to provide denser information.
Each data point of a single run is an average of 50 episode seeds, repeated three times to account for the agent's stochastic policy.
Although five samples may still be insufficient for drawing strong conclusions, we increase reliability by aggregating the trained agents' performances across both finite and endless environments.
These aggregation plots (Figures \ref{fig:finite_agg} and \ref{fig:endless_agg}) illustrate the mean and standard deviation of 15 independent seeds, allowing for a more robust statistical analysis.
The remaining sample efficiency curves, which are used for supplementary experiments (e.g., exploring TrXL's low effectiveness on endless tasks), average five runs and display standard deviations.

\subsection{Finite Environments: Transformer-XL Relies on Obs. Reconstruction}
\label{sec:finite_results}

\begin{figure}
    \centering

    \subfigure{
        \includegraphics[width=0.75\textwidth]{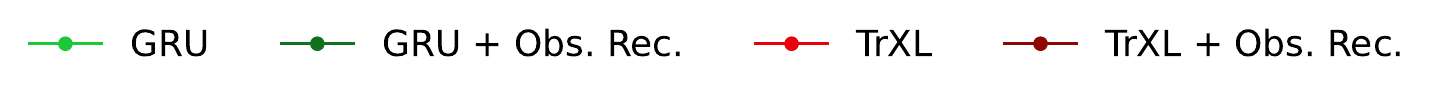}
        \label{fig:legend_flat_0}
    }
    \vspace{-0.15in}

    \setcounter{subfigure}{0}
    
    \subfigure[Mortar Mayhem]{
        \includegraphics[width=0.93\textwidth]{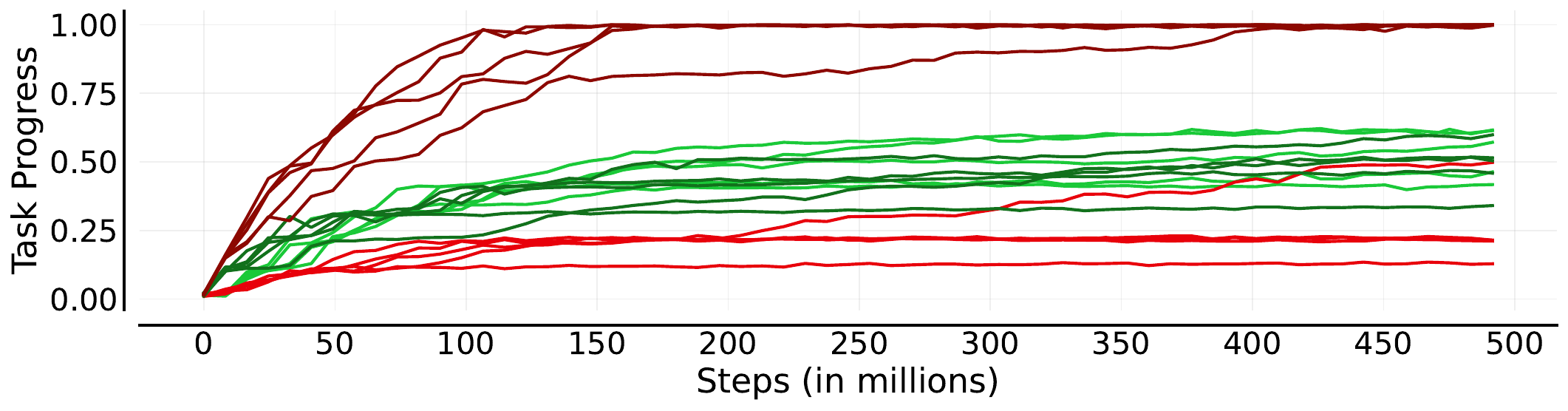}
        \label{fig:finite_mm}
    }

    \subfigure[Mystery Path]{
        \includegraphics[width=0.93\textwidth]{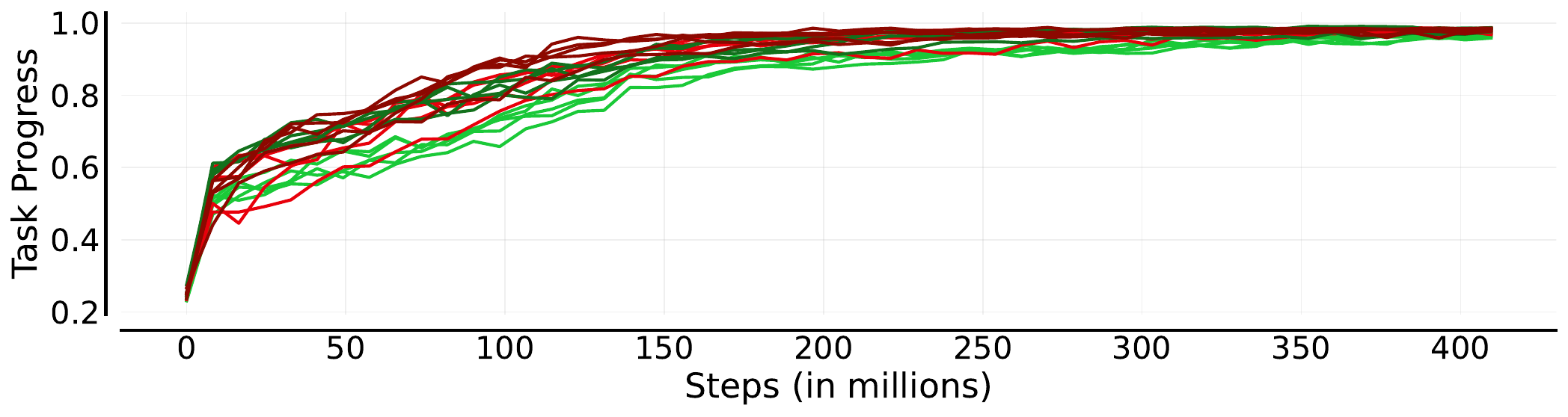}
        \label{fig:finite_mp}
    }

    \subfigure[Searing Spotlights]{
        \includegraphics[width=0.93\textwidth]{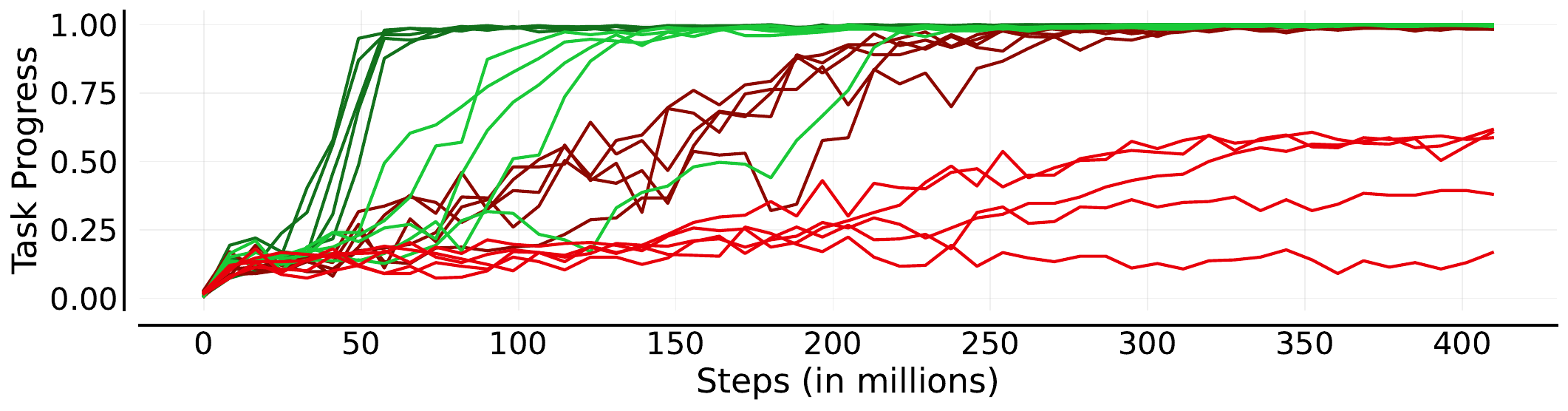}
        \label{fig:finite_ss}
    }

    \subfigure[Aggregation and Normalization Across Finite Tasks]{
        \includegraphics[width=0.93\textwidth]{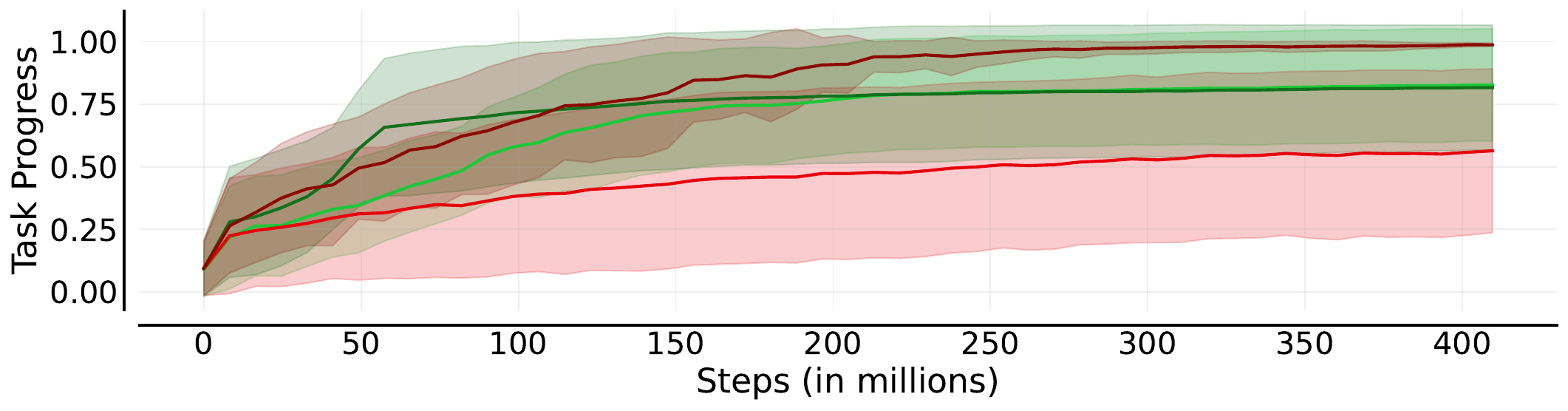}
        \label{fig:finite_agg}
    }

    \caption{Performances across Memory Gym's finite environments. For each agent, the mean performance of 5 independent seeds is shown (a, b, and c). (d) shows the mean performance and the standard deviation when normalizing and aggregating across the finite tasks.}
    \label{fig:results_finite}
\end{figure}

The introduction of the TrXL baseline motivates conducting benchmarks on the finite environments of Memory Gym, as depicted in Figure \ref{fig:results_finite}. 
Instead of comparing the success rates on the full tasks, we consider a more fine-grained task progression that provides denser information. 

In Mortar Mayhem (Figure \ref{fig:results_finite}\subref{fig:finite_mm}), the task is completed when an agent successfully executes 10 commands. 
The only agent that completes the entire task is the TrXL agent, which benefits from observation reconstruction. 
The GRU agent, which does not use this auxiliary loss, achieves an average completion of 6.2 commands at best and 4.2 commands at worst. 
TrXL without reconstruction ranges from 1.3 to 5 commands.

In Mystery Path, task progress is based on the proportion of the path completed. 
The path lengths of the evaluated episodes vary between 7 and 14 tiles. 
In this environment, all agents are effective, with the TrXL agents being more sample efficient than their GRU counterparts. 
However, when wall-time efficiency is compared, TrXL and GRU are about on par because TrXL is slower (Annex \ref{sec:wall_time}).

In Searing Spotlights (Figure \ref{fig:finite_ss}), the reported task progression measures 50\% for collecting the coin, with the remainder dedicated to successfully terminating the episode by using the exit. 
The results demonstrate that utilizing the observation reconstruction loss is key, significantly improving the effectiveness of TrXL and the sample efficiency of GRU. 
This improvement is due to rare events where the agent, the coin, and the exit are seldomly visible. 
Thus, the visual encoder benefits from the auxiliary learning signal provided by the observation reconstruction. 
Nevertheless, both GRU variants, exhibit notably higher sample efficiency compared to TrXL.

As a final assessment, we aggregate and average the normalized task progression in Figure \ref{fig:results_finite}\subref{fig:finite_agg}. 
Each curve represents the mean and standard deviation of 15 independent runs.
This statistical measure demonstrates that TrXL relies on observation reconstruction to complete 99\% of the tasks on average. 
Without it, its mean task progression drops to 56\%. 
GRU without reconstruction loss reaches a task progression of 83\%. 
When leveraging observation reconstruction, GRU appears more sample efficient, with a mean task completion of 82\%.

\subsection{Endless Environments: GRU is more effective than Transformer-XL}
\label{sec:endless_results}

\begin{figure}
    \centering

    \subfigure{
        \includegraphics[width=0.75\textwidth]{figures/results/legend_plot_flat.pdf}
        \label{fig:legend_flat_1}
    }
    \vspace{-0.15in}

    \setcounter{subfigure}{0}
    
    \subfigure[Mortar Mayhem]{
        \includegraphics[width=0.93\textwidth]{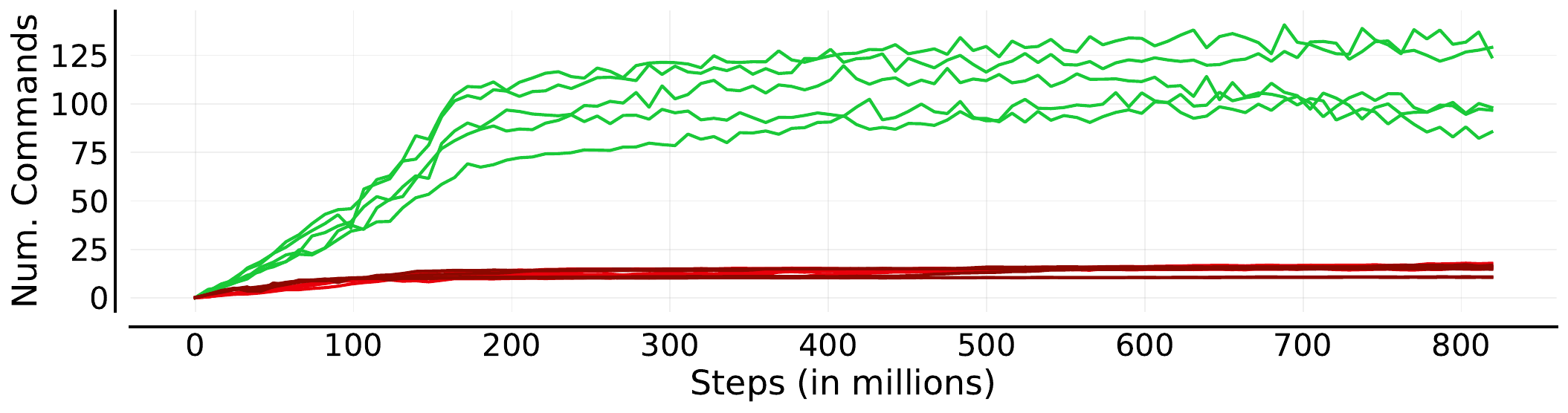}
        \label{fig:endless_mm}
    }

    \subfigure[Mystery Path]{
        \includegraphics[width=0.93\textwidth]{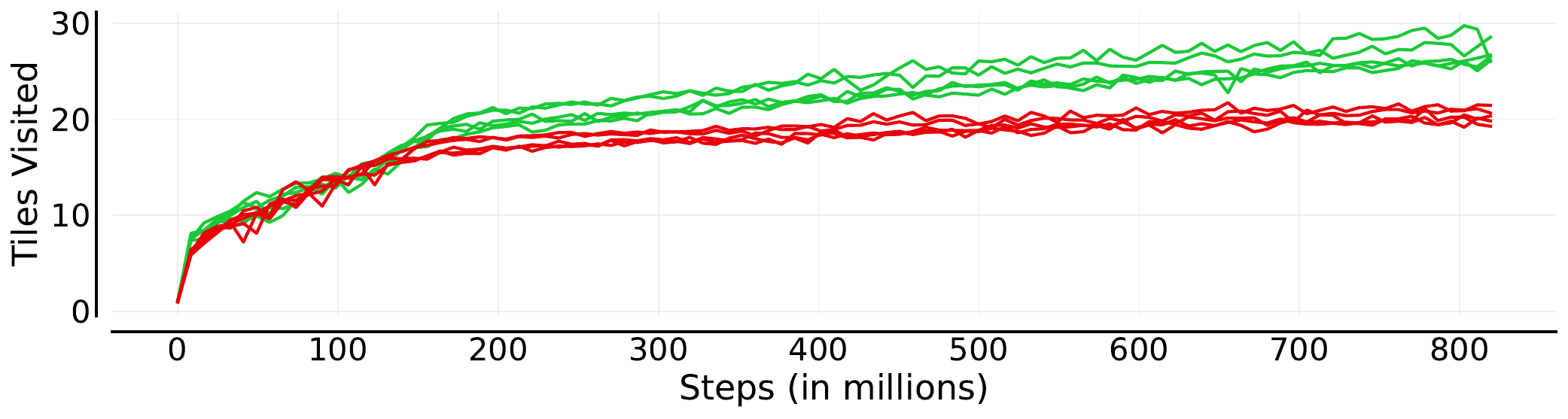}
        \label{fig:endless_mp}
    }

    \subfigure[Searing Spotlights]{
        \includegraphics[width=0.93\textwidth]{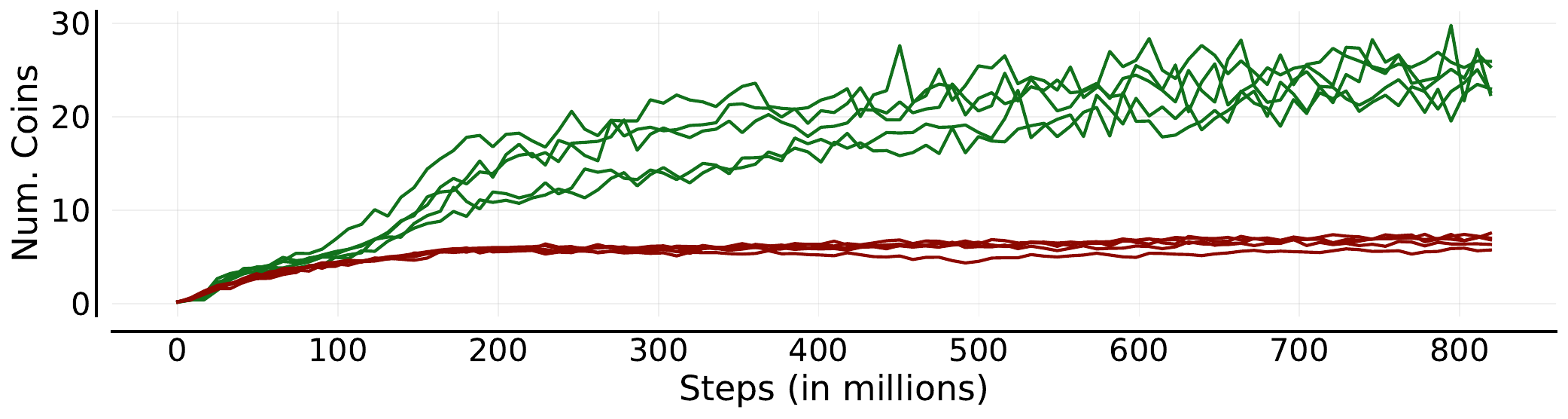}
        \label{fig:endless_ss}
    }

    \subfigure[Aggregation and Normalization Across Finite Tasks]{
        \includegraphics[width=0.93\textwidth]{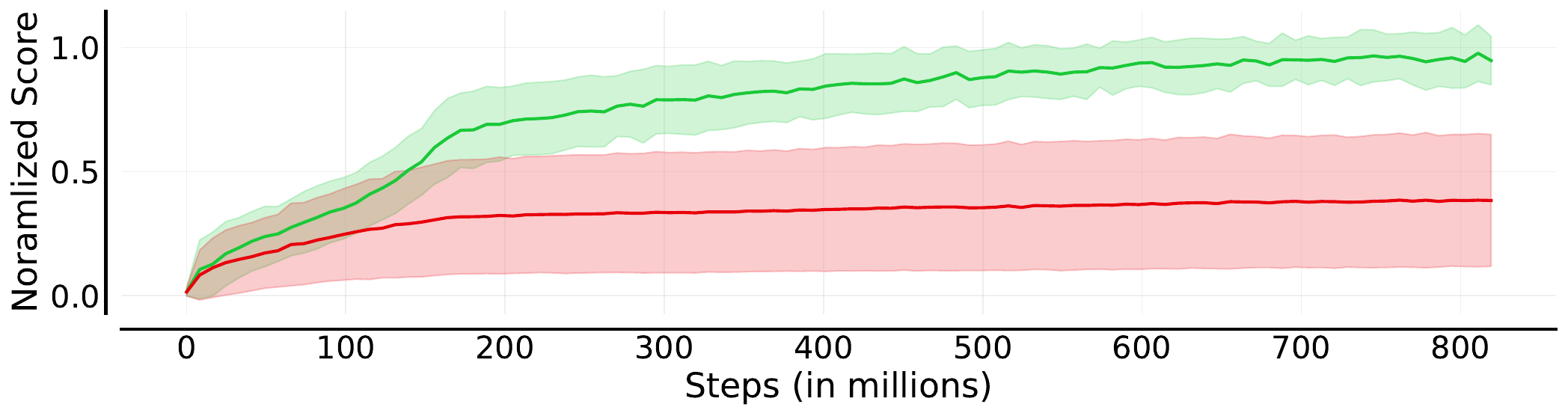}
        \label{fig:endless_agg}
    }

    \caption{GRU's consistent superiority over TrXL is revealed in Memory Gym's endless environments. For each agent, the mean performance of 5 independent seeds is shown (a, b, and c). (d) shows the mean performance and the standard deviation when normalizing and aggregating across the endless tasks.}
    \label{fig:endless_results}
\end{figure}

To our surprise, across all three endless environments, the recurrent agent consistently proves to be much more effective than the transformer-based agent.
This diverges from the observations made by \cite{parisotto2020} and \cite{Lampinen2021htmBallet}, where an LSTM was found to be less effective.
This subsection presents the raw results, and the following one discusses how finite tasks provide only a truncated view of the memory capabilities of the evaluated agents.

Notably, the most substantial gap between GRU and TrXL emerges in the results of Endless Mortar Mayhem (Figure \ref{fig:endless_mm}). 
In this context, GRU attains an impressive range of 84 to 120 executed commands using roughly 300 million steps, whereas both TrXL baselines only attain an average of 17 commands. 
The best data point for the recurrent agent measures 140 successfully executed commands on average for a single training run. 
When taking the mean of the 5 GRU runs, the best point in time scores 115 commands with a standard deviation of 15.

The outcomes depicted in Figure \ref{fig:endless_mp} illustrate the results obtained from the Endless Mystery Path environment. 
In this case, GRU remains more effective than TrXL, although the gap between their performances is narrower compared to Endless Mortar Mayhem. 
In this setting, the best TrXL run visits 21 tiles of the path, while GRU visits 29 tiles at best. 
After roughly 150 million steps, all GRU curves continuously distance themselves from TrXL. 
Upon training completion, the mean of all GRU runs is 26 tiles, while the TrXL ones average 20 tiles.

In Endless Searing Spotlights (Figure \ref{fig:endless_ss}), both agents utilize the observation reconstruction loss.
TrXL is capable of collecting 7.5 coins at maximum, while GRU's best performance is nearly 4 times higher with 29.8 collected coins.
The final performance after training measures 23.9 coins for GRU and 6.7 ones for TrXL.

All results obtained from the endless environments show a clear performance gap between GRU and TrXL. 
This is also salient in the aggregation plot of Figure \ref{fig:endless_agg}, which combines all GRU and TrXL runs, including those from Endless Searing Spotlights that use observation reconstruction. 
Hence, both the green and red curves aggregate 15 independent seeds. 
The data is normalized by the maximum mean score of the GRU agent in each environment. 
Figure \ref{fig:endless_agg} shows that the GRU agent peaks at 0.98, while TrXL reaches a normalized score of 0.39. 
Additionally, the error bars of the standard deviations do not overlap after roughly 250 million steps.

\subsection{Memory Effectiveness: Finite Boundaries vs. Endless Potential}
\label{sec:results_effectiveness_efficiency}

In this subsection, we highlight that the insights derived from finite environments provide only a truncated view of the true memory capabilities of the benchmarked agents.
To recall, higher memory effectiveness implies the ability to memorize more information over extended periods.

Mortar Mayhem stands as an exception among finite environments, where varied levels of effectiveness are observable. 
One might assume that these differences would align with the results from endless tasks, yet this alignment does not occur. 
Notably, in the endless setting, GRU completes significantly more commands than the 6.2 observed in the finite environment. 
This increased quantity may be attributed to two key factors. 
Firstly, the environment's curriculum begins at an easier level by only demanding one command initially. 
Secondly, Endless Mortar Mayhem allows for two strategies for memorizing commands: the direct visualization of the command and the agent’s ability to infer commands based on its past actions. 
This may allow for the agent's policy to adapt faster to growing difficulties.
Regarding episode lengths, when completing 10 commands in the finite tasks, episodes last for 274 steps. 
In the endless task, GRU plays for more than 3000 steps and TrXL for more than 400 steps.

In the Mystery Path environment, challenges are limited to a maximum path length of 14 tiles, providing only a basic indication of the memory requirements.
This measure is approximate, as the procedurally generated path may include multiple turns, increasing the likelihood of the agent falling off the path.
Both GRU and TrXL effectively master this challenge, but they surpass these capabilities in the endless version.
While successful episodes in the finite task last for about 100 steps, GRU reaches more than 600 steps on average in the endless task, while TrXL approaches 400 steps.

Quantifying performance in Searing Spotlights involves navigating to two entities (coin and exit), while the agent must maintain its hidden position.
In this finite task, effective episodes typically last fewer than 50 steps.
However, in Endless Searing Spotlights, GRU and TrXL can gather more coins and maintain their hidden positions for 691 and 240 steps on average, respectively.
This performance demonstrates that both agents not only achieve higher scores but also sustain their concealed positions for longer durations than observed in the finite version of Searing Spotlights.

To conclude, endless tasks demonstrate their potential to reveal more effective memory capabilities in both baselines beyond what is evident in finite tasks, as more information needs to be memorized and maintained over prolonged time horizons.

To conclude, finite environments provide only a truncated view of the agents' capabilities. 
Instead, the endless environments demonstrate their potential to reveal more effective memory capabilities in both baselines beyond what is evident in finite tasks, as more information needs to be memorized and maintained over ever-growing time horizons.

\subsection{Investigating Transformer-XL's Surprisingly Low Effectiveness}

\begin{figure}[t!]
    \centering

    \subfigure{
        \includegraphics[width=0.99\textwidth]{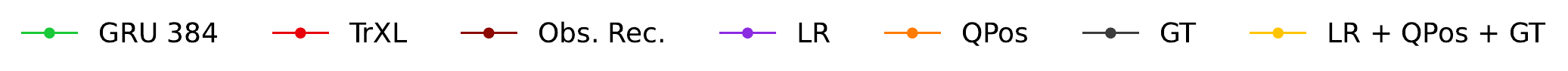}
        \label{fig:legend_flat_emm}
    }
    \vspace{-0.25in}
    
    \subfigure{
        \includegraphics[width=0.95\textwidth]{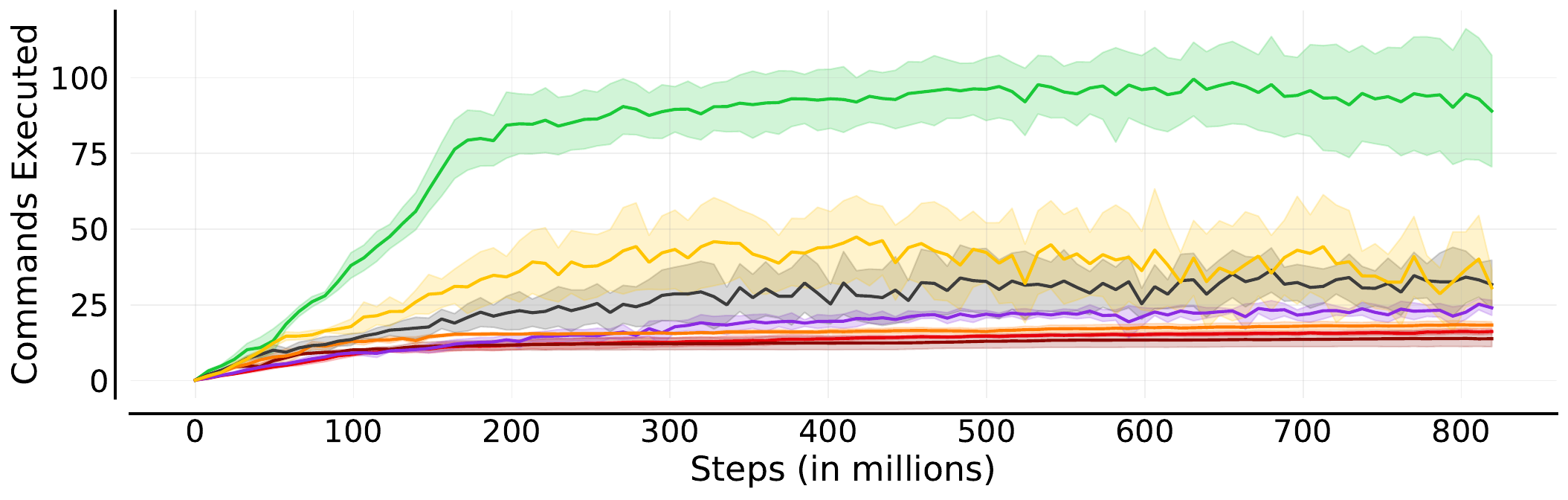}
        \label{fig:emm_trxl_novel}
    }
    \caption{Experiments on Endless Mortar Mayhem with varied TrXL configurations.
    The learning rate (LR) is adjusted to decay from $2.75$e-$4$ to $1.0$e-$4$ over 160 million steps, compared to the previous range from $2.75$e-$4$ to $1.0$e-$5$.
    We also incorporate observation reconstruction (Obs. Rec.) and observe the greatest improvement based on a ground truth estimation head (GT) as sanity check.
    Another test augments TrXL's query with absolute positional encoding (QPos).
    The final experiment combines the optimized learning rate, augmented query, and ground truth estimation.}
    \label{fig:emm_trxl}
\end{figure}

The unexpectedly low performance of TrXL prompts us to examine several hypotheses.
Endless Mortar Mayhem, presenting the largest performance gap, serves as our primary environment for these investigations.

\subsubsection{Inadequate Networks Capacity}

Our TrXL baseline comprises 2.8 million trainable parameters, while GRU consists of 4.05 million, prompting the question of whether TrXL's model architecture lacks the necessary capacity.
To investigate this, we conducted experiments varying the number of layers (2, 3, 4), the embedding dimension size (256, 384, 512), and the memory window length (256, 384, 512).
None of these adjustments led to closing the performance gap to GRU.
Even when the GRU cell is scaled down to a dimension of 384, totaling 2.7 million parameters, it still surpasses TrXL by a large margin (Figure \ref{fig:emm_trxl}).
Note that TrXL's maximum context length, defined by the number of layers and window length (Equation \ref{eq:length}), reaches 766 timesteps with 3 layers and a window length of 256.
Therefore, the context length is not the limiting factor in any endless environment.

Scaling up TrXL also increases its demand for GPU memory.
When scaling up multiple architecture details, our implementation and hyperparameters may cause the training to exceed the available 40GB GPU memory of an NVIDIA A100.
Workarounds include reducing the batch size or transferring training data between the GPU and CPU, but these options significantly worsen wall-time efficiency.
Furthermore, another indication of untapped capacity lies in the amplification of the learning signal, which we elaborate on in the following section.

\subsubsection{Weak Learning Signal}
\label{sec:weak_signal}

Figure \ref{fig:emm_trxl} depicts several additional experiments aimed at amplifying the learning signal. 
We made a naive adjustment to the learning rate schedule, utilized observation reconstruction, and introduced a ground truth estimation head. 
This ground truth estimation head is added to the actor-critic model architecture and predicts the next target position to move to. 
Labels are provided by the environment, and the estimates are optimized using the mean-squared error loss.
While ground truth information is typically unavailable and considered inappropriate for this benchmark, it serves as a useful sanity check, helping to determine if an additional learning signal is advantageous in this context.

When training incorporates ground truth estimation, there is improvement in the agent's policy. 
Previously, the mean number of completed commands stood at 16; with ground truth estimation, it reaches a mean of up to 36 commands. 
As a default and during the initial 160 million training steps, the learning rate linearly decays from $2.75 \times 1.0$e-$4$ to $1.0$e-$5$. 
By setting the final decayed learning rate to $1.0$e-$4$, the mean performance hits 25 commands. 
The results of leveraging ground truth estimation or tuning the learning rate schedule imply that the model's current capacity is capable of retaining more information.

\subsubsection{Lack of Temporal Information in the Initial Query}

The next hypothesis pertains to the initial query of the first TrXL layer.
This query is constructed solely based on the features extracted from the observation encoders at the current time step, encompassing only minimal temporal information.
In contrast, the query of the second layer draws from the aggregated outcome of the memory window, thus capturing more substantial temporal information.
However, we believe that the initial query could be further enriched with additional information.
In Figure \ref{fig:emm_trxl}, one experiment augments the query with absolute positional encoding, providing direct access to the current time step.
Despite this enhancement, the agent's performance only moderately improves, reaching a mean of 18 completed commands.
We also explored embedding the original query with the previous output of the TrXL block.
However, this simple approach did not yield positive results.

None of the aforementioned measures individually reach the performance level of the GRU agent. 
However, when combined, they achieve a mean performance of 47 executed commands at best.

\subsubsection{Harmful Off-Policy Data in the Episodic Memory}


Next, it can be discussed whether the combination of PPO and the cached hidden states of TrXL's episodic memory pose a threat.
PPO is an on-policy algorithm that consumes the training data for a few epochs.
Once the second training epoch on the same batch commences, the batch is already considered off-policy.
However, PPO's loss function can mitigate this issue and hence allows for several epochs.
But this mitigation is likely limited and we wonder whether this issue is enlarged by the cached hidden states.
In our current training setup, the agent collects 512 steps of data.
If an episode exceeds 512 steps and is still running, the older hidden states from that episode become outdated over multiple iterations of PPO.
To address this, future research should investigate the impact of stale hidden states on training efficacy or how one can solve the issue of truncated episodes during optimization.
One potential strategy could involve computing more fresh hidden states alongside utilizing cached ones.
Alternatively, cached hidden states could be periodically recomputed, albeit at the expense of increased inference overhead.
Drawing parallels, \cite{kapturowski2018recurrent} explored similar challenges with their recurrent, off-policy algorithm R2D2.

\subsubsection{Indistinguishable Absolute Positional Encoding}

The final hypothesis concerns a potential limitation of the absolute positional encoding.
Without positional encoding, the agent cannot differentiate the time between past observations.
As a default, we rely on absolute positional encodings.
This encoding always spans across the potential maximum episode length.
In the finite environments, the longest range is set to 512, being the sequence length, which is also used in the original transformer \citep{Vaswani2017attention}.
For the endless tasks, this encoding is set up for 2048 steps, an episode duration that is yet unreachable to the TrXL agents.

The original work of TrXL introduces a relative positional encoding \citep{Dai2019TrXL} as a consequence of extended context lengths.
However, when they refer to the absolute positional encoding, they only apply it to the range of the current window.
If a window has a length of 4, the positions within the window are always the same (i.e. $[0, 1, 2, 3]$), even though the window moves across the entire sequence.
In our case, the absolute positional encoding considers the entire sequence.
If the window has progressed in time, the applied positional encoding relies on the current time step (i.e. $[t-3, t-2, t-1, t]$).

Nevertheless, the absolute positional encoding needs to be defined prior training.
When scaling it up, the question arises whether the agent's model is capable of differentiating time steps, because a longer length indicates a more fine-grained sample frequency over the underlying positional encoding.
To test this, we trained a TrXL agent on the finite task of Mortar Mayhem and configured an absolute positional encoding of length 1024 and 2048.
Both experiments, which were repeated twice, resulted in an agent that poorly completes only 1 to 2 commands.
This demands for questioning the absolute positional encoding in endless tasks.
But if the elements of the absolute position encoding of length 2048 are not distinguishable, how do those agents master more than 10 commands in Endless Mortar Mayhem?

\begin{figure}
    \centering

    \subfigure{
        \includegraphics[width=0.5\textwidth]{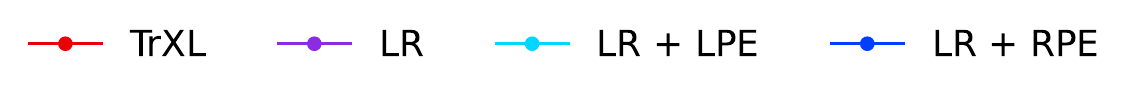}
        \label{fig:legend_flat_emm_pe}
    }
    \vspace{-0.25in}
    
    \subfigure{
        \includegraphics[width=0.95\textwidth]{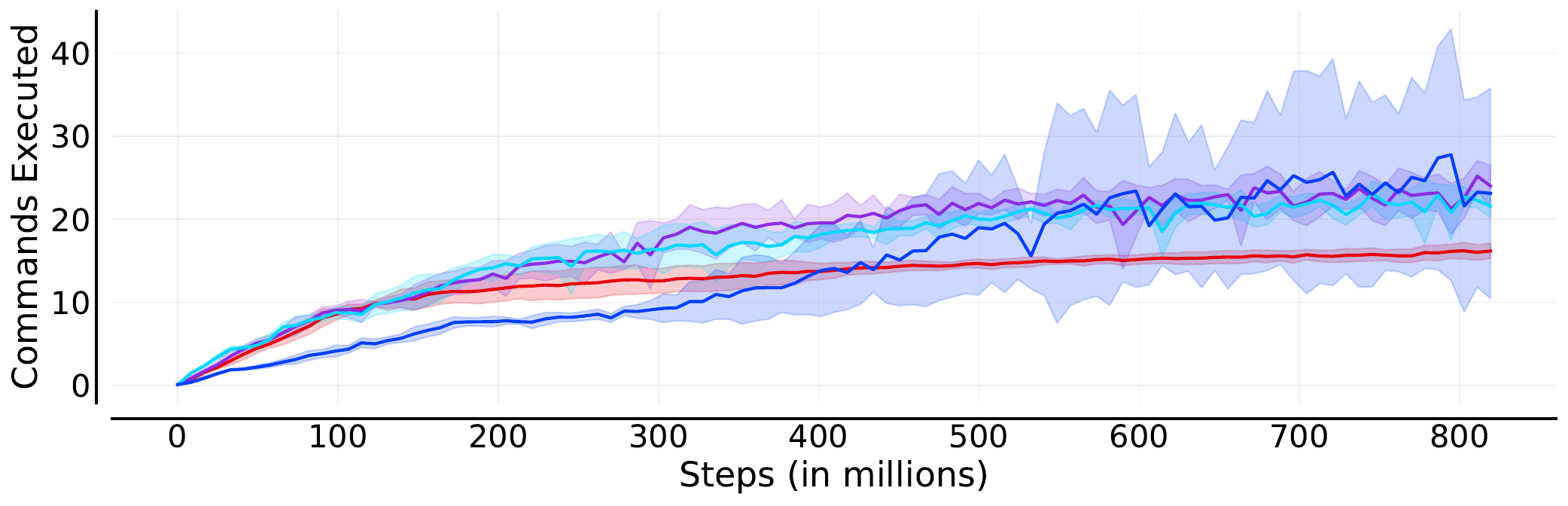}
        \label{fig:emm_trxl_novel_pe}
    }
    \caption{Experiments on Endless Mortar Mayhem with varied positional encoding.
    The learning rate schedule (LR) is adjusted to decay from $2.75$e-$4$ to $1.0$e-$4$ over 160 million steps, compared to the previous $1.0$e-$5$.
    The acronym LPE refers to a learned positional encoding, while RPE depicts the relative one.}
    \label{fig:emm_trxl_pe}
\end{figure}

If this assumption is correct---that the absolute positional encoding is the culprit in our TrXL baseline---leveraging learned or relative positional encodings instead should improve performance.
As seen in Figure \ref{fig:emm_trxl_pe}, no notable advancement is accomplished by both variants.
We observed two outliers among the 5 runs that utilized relative positional encoding.
One underperformed, completing only 16 commands on average, while the other excelled, finishing 56 commands on average.
Given this discussion, a more in-depth exploration of positional encoding in this context is advisable.

\subsection{Recurrence is not Vulnerable to Spotlight Perturbations}

Training recurrent agents in Searing Spotlights with previously established hyperparameters (refer to the last column of Table \ref{tab:hyperparameters}) does not yield a meaningful policy.
Despite numerous attempts at hyperparameter tuning, architecture adjustments, and task modifications, Searing Spotlights remained as a puzzle.
Even with full observability, where all entities are permanently visible, the performance is catastrophic.
These challenges prompted the hypothesis that spotlights might introduce noise, heavily affecting the agent's policy \citep{pleines2023memory}.
Support for this theory came from experiments in a simplified BossFight environment \citep{ProcgenCobbe202}, using identical spotlight dynamics as in Searing Spotlights, suggesting a potential vulnerability of recurrent agents to such perturbation \citep{pleines2023memory}.

\input{tables/searing_spotlights_study}

However, the results presented in Figure \ref{fig:finite_ss} neglect this hypothesis.
This prompted us to conduct an ablation study, beginning with the previous hyperparameters.
In this study, only a single hyperparameter or environment option was replaced with the current one at a time.
Table \ref{tab:ss_ablation} presents the results for Searing Spotlights when full observability is ensured.
Initially, observation reconstruction was presumed to be the solution.
However, the crucial adjustment proved to be the decision to avoid normalizing advantage estimates during optimization.
Only the GRU agent, which omits advantage normalization, and the memory-less agent, based entirely on previous hyperparameters, recorded noteworthy success rates of 100\% and 96\%, respectively.

A salient observation emerges when tracking the gradient norms across the agent models during training.
With normalized advantages, the gradient norm is typically more than 12 times greater than when normalization is bypassed.
This suggests that during gradient descent, excessively large optimization steps might be taken, resulting in an oscillating behavior that fails to converge to a local minimum.
Simultaneously, unnormalized advantages remain at a scale of $2$e-$4$, but post-normalization, the magnitudes stretch beyond $5$.
Considering that the policy ratio  is modulated by these advantage estimates \citep{Schulman2017}, normalization introduces a strikingly different scale than its absence.
Standard practice typically normalizes advantages by subtracting the mean and dividing by the standard deviation, often applied at the mini-batch level.

Several comprehensive studies have delved into such nuances of PPO's implementation \citep{Engstrom2020,andrychowicz2021what,huang2022ppo}.
Among them, only \cite{andrychowicz2021what} turned off advantage normalization and found negligible performance impact.
However, their extensive study focused on control tasks with undiscounted returns in the thousands, contrasting Memory Gym's finite environments where the undiscounted returns do not exceed 2.
During hyperparameter optimization, turning off advantage normalization also enhanced performance in Mortar Mayhem and Mystery Path.
Given the critical importance of utilizing raw advantages, there remains an opportunity to delve even deeper into PPO to comprehend and refine its essential aspects.

%% file: tables/searing_spotlights_study.tex
\begin{table}[t!]
\caption{Results from the ablation study on Searing Spotlights with full observability. Each experiment was run 5 times, though the data sourced directly from the training process deviates from our standard protocol. Every PPO iteration (of 5,000 total) produces a data point that reflects the average over the last 100 episodes. Rows are ordered in descending fashion by the mean success rate. The mean is computed from the top-performing data point, averaged across repetitions, from the specific experiment. Consequently, the time point is represented as a percentage duration. The gradient norm is accumulated over all data points within the given duration.}
\label{tab:ss_ablation}
\vspace{0.1in}
\arrayrulecolor{black}

\begin{tabular}{|l|rr|rr|r|rr|}
\hline
\rowcolor[HTML]{C0C0C0} 
\textbf{Experiment} & \multicolumn{2}{c|}{\cellcolor[HTML]{C0C0C0}\textbf{Value}}                 & \multicolumn{2}{c|}{\cellcolor[HTML]{C0C0C0}\textbf{Success Rate}}           & \multicolumn{1}{c|}{\cellcolor[HTML]{C0C0C0}\textbf{Duration}} & \multicolumn{2}{l|}{\cellcolor[HTML]{C0C0C0}\textbf{Gradient Norm}}          \\ \hline
\textbf{}           & \multicolumn{1}{c|}{New}                         & \multicolumn{1}{c|}{Old} & \multicolumn{1}{c|}{Mean}                         & \multicolumn{1}{c|}{Std} & \multicolumn{1}{c|}{}                                          & \multicolumn{1}{c|}{Mean}                         & \multicolumn{1}{c|}{Std} \\ \hline
\rowcolor[HTML]{DAE8FC} 
Advantage Norm.     & \multicolumn{1}{r|}{Off} & On                       & \multicolumn{1}{r|}{1.00} & 0.00                     & 31\%                                                           & \multicolumn{1}{r|}{0.02} & 0.01                     \\ \hline
No Memory           & \multicolumn{1}{r|}{}                            &                          & \multicolumn{1}{r|}{0.96}                         & 0.11                     & 99\%                                                           & \multicolumn{1}{r|}{0.28}                         & 0.03                     \\ \hline
\rowcolor[HTML]{EFEFEF} 
Agent Health Points & \multicolumn{1}{r|}{5}   & 10                       & \multicolumn{1}{r|}{0.79} & 0.35                     & 99\%                                                           & \multicolumn{1}{r|}{0.30} & 0.02                     \\ \hline
Transformer-XL      & \multicolumn{1}{r|}{}                            &                          & \multicolumn{1}{r|}{0.69}                         & 0.27                     & 99\%                                                           & \multicolumn{1}{r|}{0.30}                         & 0.03                     \\ \hline
\rowcolor[HTML]{EFEFEF} 
Agent Speed         & \multicolumn{1}{r|}{3.0} & 2.5                      & \multicolumn{1}{r|}{0.58} & 0.22                     & 65\%                                                           & \multicolumn{1}{r|}{0.33} & 0.06                     \\ \hline
Clip Range          & \multicolumn{1}{r|}{0.1}                         & 0.2                      & \multicolumn{1}{r|}{0.52}                         & 0.25                     & 99\%                                                           & \multicolumn{1}{r|}{0.27}                         & 0.05                     \\ \hline
\rowcolor[HTML]{EFEFEF} 
Max Episode Length  & \multicolumn{1}{r|}{256} & 512                      & \multicolumn{1}{r|}{0.48} & 0.27                     & 99\%                                                           & \multicolumn{1}{r|}{0.37} & 0.04                     \\ \hline
Old Hyperparameters & \multicolumn{1}{r|}{}                            &                          & \multicolumn{1}{r|}{0.43}                         & 0.06                     & 96\%                                                           & \multicolumn{1}{r|}{0.38}                         & 0.06                     \\ \hline
\rowcolor[HTML]{EFEFEF} 
Observation Rec.    & \multicolumn{1}{r|}{On}  & Off                      & \multicolumn{1}{r|}{0.39} & 0.06                     & 96\%                                                           & \multicolumn{1}{r|}{0.40} & 0.04                     \\ \hline
Max Gradient Norm   & \multicolumn{1}{r|}{0.25}                        & 0.5                      & \multicolumn{1}{r|}{0.36}                         & 0.05                     & 99\%                                                           & \multicolumn{1}{r|}{0.24}                         & 0.01                     \\ \hline
\end{tabular}
\end{table}

%% file: content/conclusion.tex
\section{Conclusion}

In this study, we advanced Memory Gym's environments to novel endless tasks, tailored to benchmark memory-based Deep Reinforcement Learning algorithms.
These tasks feature a mounting challenge similar to the cumulative memory game ``I packed my bag".
As the agent's policy refines, the task dynamically expands, serving as an automatic curriculum.
This innovative framework enables a thorough evaluation of memory-based agents, emphasizing their overall effectiveness beyond mere interaction efficiency with the environment.
Our experimental results reveal that agents trained on these endless tasks consistently show stronger capabilities than those trained on finite tasks, demonstrating that finite environments offer only a truncated view on an agent's full memory capabilities.

To foster a more inclusive research landscape, we contributed an open-source PPO baseline powered by Transformer-XL (TrXL).
This baseline employs an attention mechanism applied to an episodic memory with a sliding window.
When benchmarked against Memory Gym, the TrXL baseline competes head-to-head with another prominent baseline found in Gated Recurrent Unit (GRU).
Notably, in finite environments, TrXL is more effective than GRU, but strongly depends on the observation reconstruction loss.
Our most unexpected revelation is the comeback of GRU in the endless environments.
GRU surpasses TrXL by large margins, while also being computationally more efficient.
Further probing into the Endless Mortar Mayhem environment revealed only marginal improvements for TrXL upon enriching the learning signal.

This prompts future investigations into potential reasons behind TrXL's limitations, such as the possibility of episodic memory staleness or an initial query lacking temporal awareness.
As we move forward, it will be compelling to discern the performance thresholds of other memory mechanisms and DRL algorithms.
In addition to examining more attention-based and recurrent architectures, it may be worth exploring structured state space sequence models \citep{Gu2022S4D}, given their recent debut in DRL \citep{lu2023structured}.


%% file: content/annex.tex
\section{Environment Parameters}
\label{sec:environment_parameters}

\input{tables/env_parameters}

\input{tables/env_parameters_endless}

\newpage

\section{Hyperparameters}
\label{sec:hyperparameters}

\input{tables/hyperparameters}

Table \ref{tab:hyperparameters} presents the hyperparameters utilized in our final experiments unless otherwise specified.
The learning rate and entropy coefficient undergo linear decay from their initial to final values.
This decay occurs exclusively during the first 10,000 PPO updates (equivalent to 163,840,000 steps).
As training progresses beyond this point, the final hyperparameter serves as a lower threshold.
Regarding the sequence length and memory window length, they are determined based on the longest possible episode in the finite environments.
In the case of endless environments, the so far best sequence length is fixed at 512 for GRU and 256 for TrXL.

Our hyperparameter search was conducted using a customized implementation built upon the optuna tuning framework \citep{Akiba2019optuna}.
The objective was to discover a set of hyperparameters that demonstrate strong performance across Memory Gym's environments, for the GRU and TrXL baselines individually.
Considering limited resources and the computational expense of tuning, we restricted the search space to a limited number of options.
The majority of tuning experiments focused on the finite environments.
Therefore, we do not claim to provide optimal hyperparameters.

For each environment, we established individual tuning experiments corresponding to the choices presented in Table \ref{tab:hyperparameters}.
Each experiment, or trial, involved a baseline set of hyperparameters that was modified by selecting a single parameter choice from the entire search space.
Notably, we did not train permutations of all available choices at this stage.
To optimize resource utilization, we implemented optuna's percentile pruner, which retained the top 25 percentile of trials.
Trials that performed below this percentile were pruned, but only after surpassing 2,000 PPO updates as a warm-up phase.

After completing the single experiments, we expanded the hyperparameter search by allowing for permutations sampled using optuna's Tree-structured Parzen Estimator.
It is important to note that individual trials were not repeated, but since tuning was conducted across multiple environments, one could consider them as repetitions in a broader sense.

The batch size is determined by multiplying the number of environment workers by the number of worker steps performed by each worker to gather training data.
To efficiently utilize the resources of a system with 32 CPU cores and 40GB VRAM on an nVidia A100 GPU, we chose a batch size of 16,384 samples.
However, when scaling up the model, it is possible to exceed the available GPU memory.
In such cases, there are two options: reducing the batch size or outsourcing the data to CPU memory. 
Both of these workarounds come with increased expenses in terms of wall-time.

\section{Results of Minor Baselines on the Finite Environments}
\label{sec:minor_baselines}

\begin{figure}
    \centering
    \subfigure[Mortar Mayhem Act Grid]{\includegraphics[clip, trim=0 1.4cm 0 0, width=0.33\textwidth]{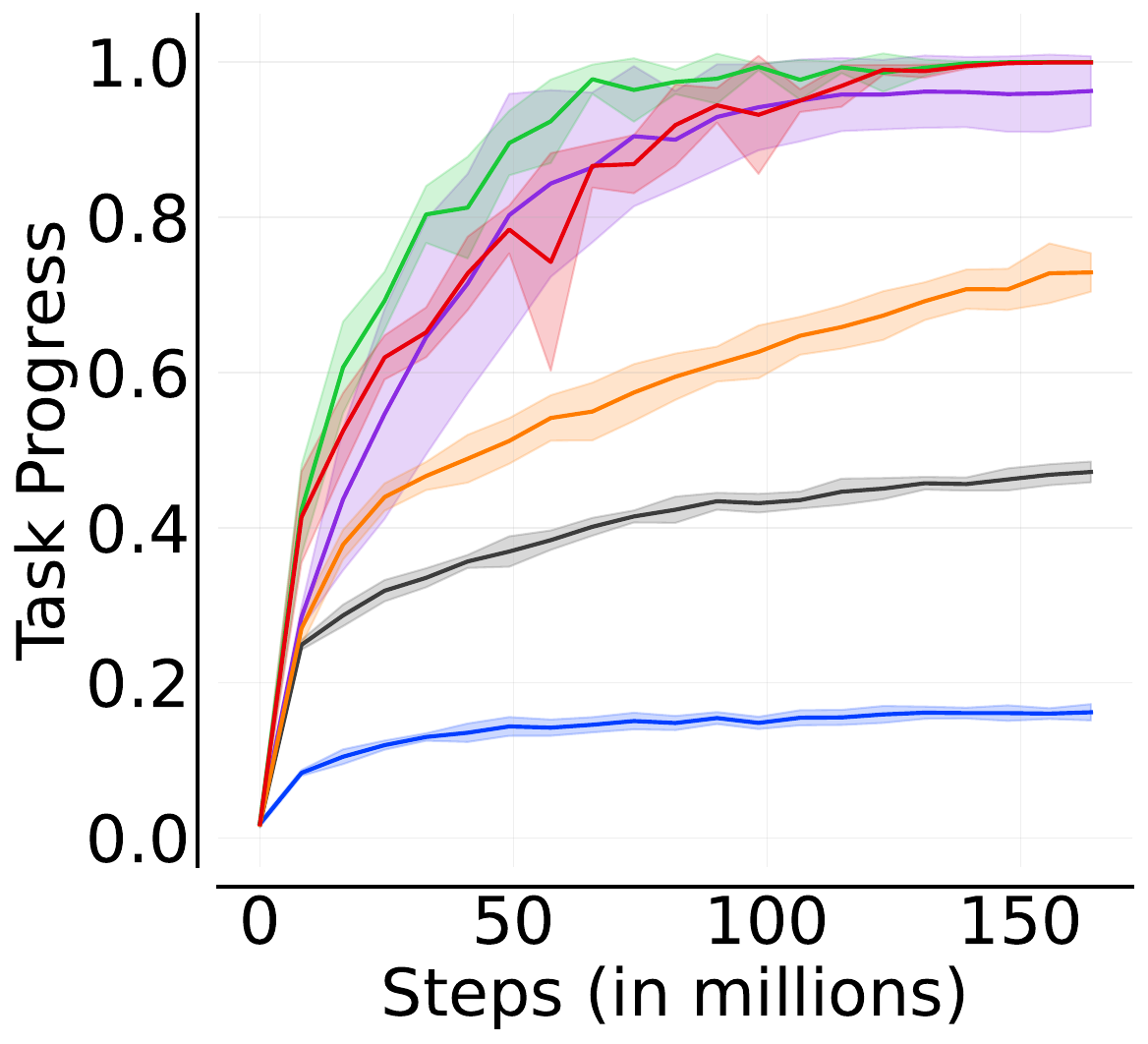}\label{fig:res_mm_act_grid_2}}
      \hfill
    \subfigure[Mortar Mayhem Grid]{\includegraphics[clip, trim=1.3cm 1.4cm 0 0, width=0.31\textwidth]{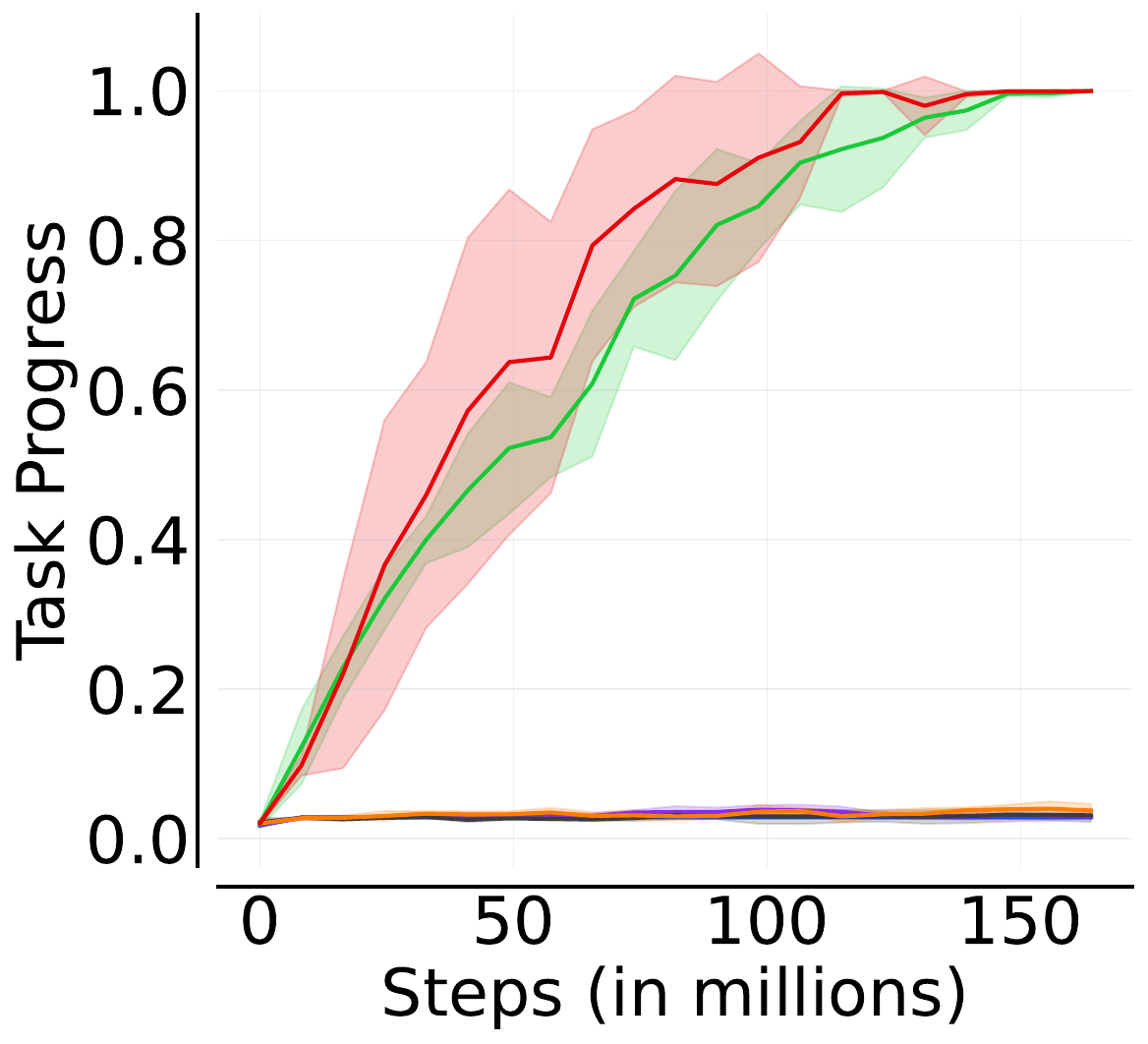}
    \label{fig:res_mm_grid_2}}
    \hfill
    \subfigure{\raisebox{0.0cm}{\includegraphics[width=0.325\textwidth]{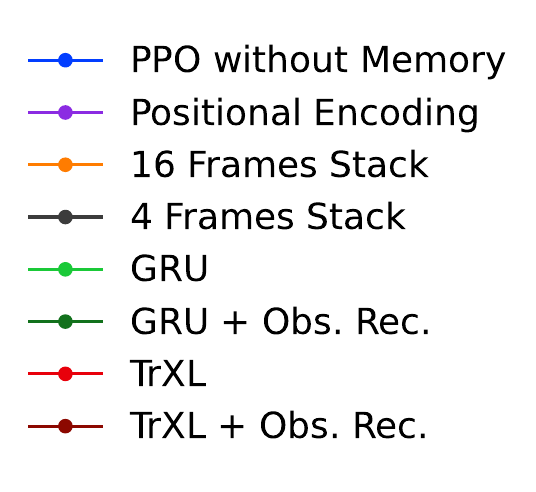}}\label{fig:res_legend_2}}\hfill

    \subfigure[Mystery Path Grid (Cues On)]{\includegraphics[width=0.33\textwidth]{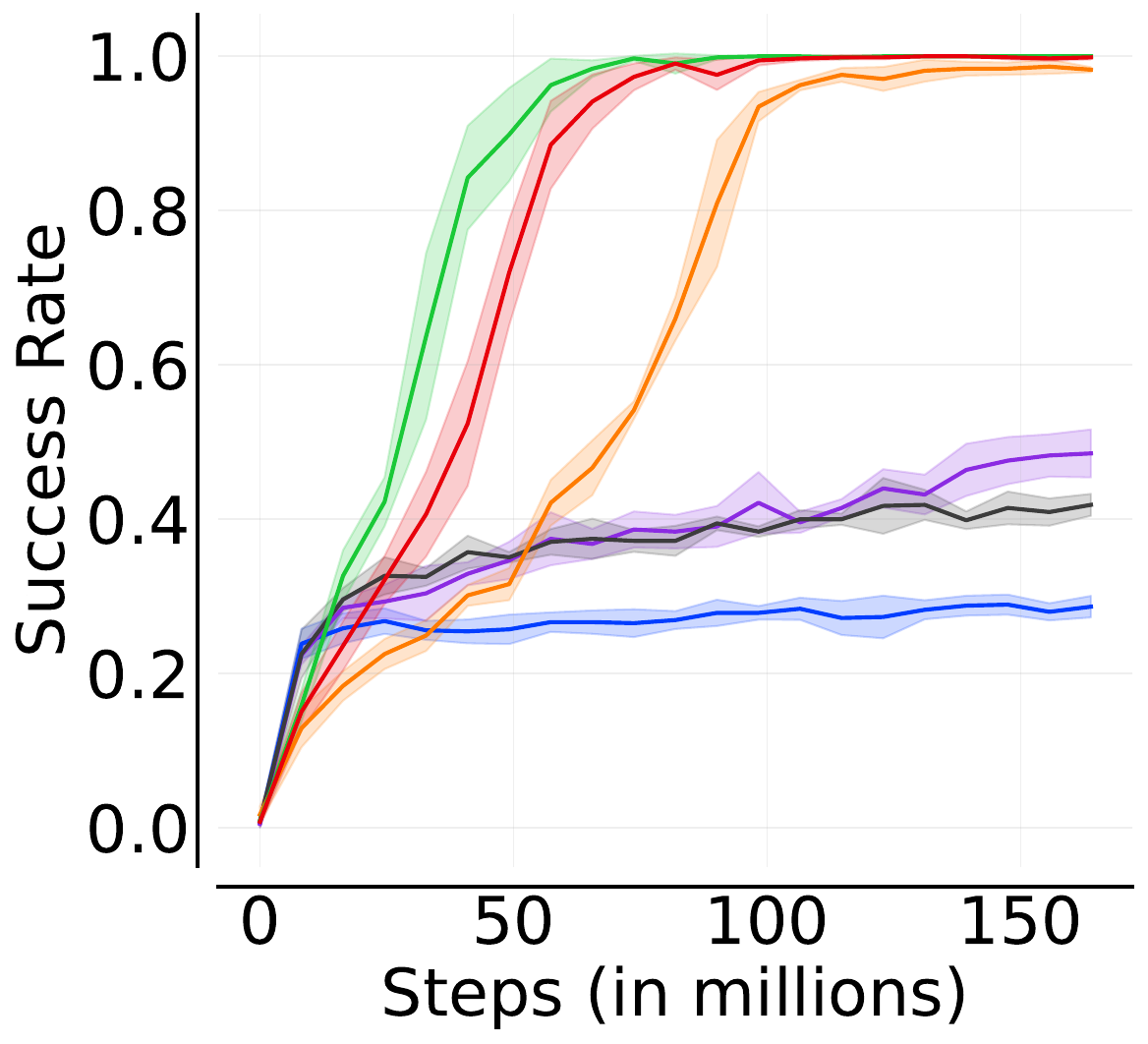}\label{fig:res_mp_grid_cues_2}}
    \hfill
    \subfigure[Mystery Path Grid]{\includegraphics[clip, trim=1.3cm 0 0 0,width=0.31\textwidth]{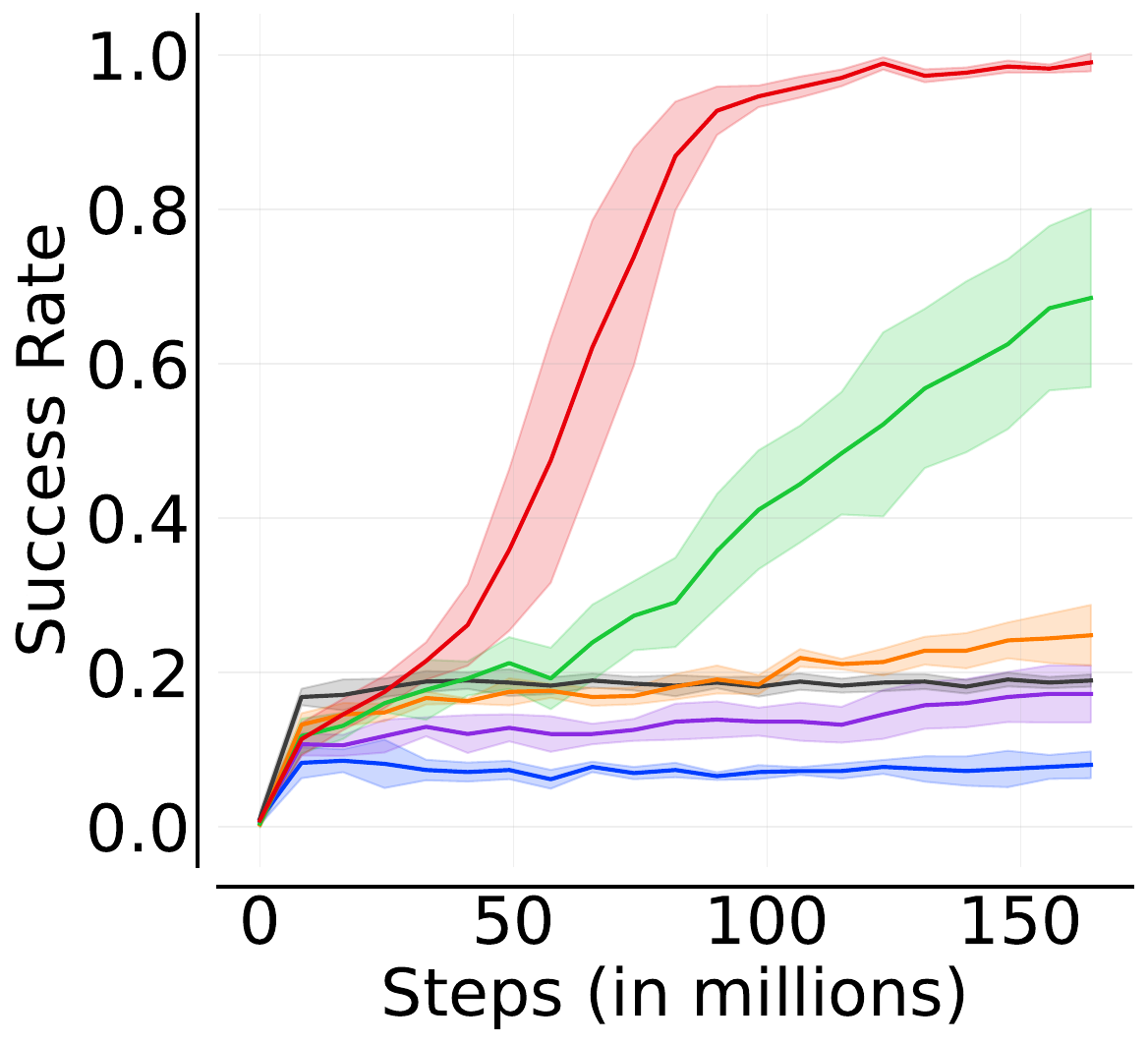}\label{fig:res_mp_grid_2}}
    \hfill
    \subfigure[Searing Spotlights]{\includegraphics[clip, trim=1.3cm 0 0 0,width=0.31\textwidth]{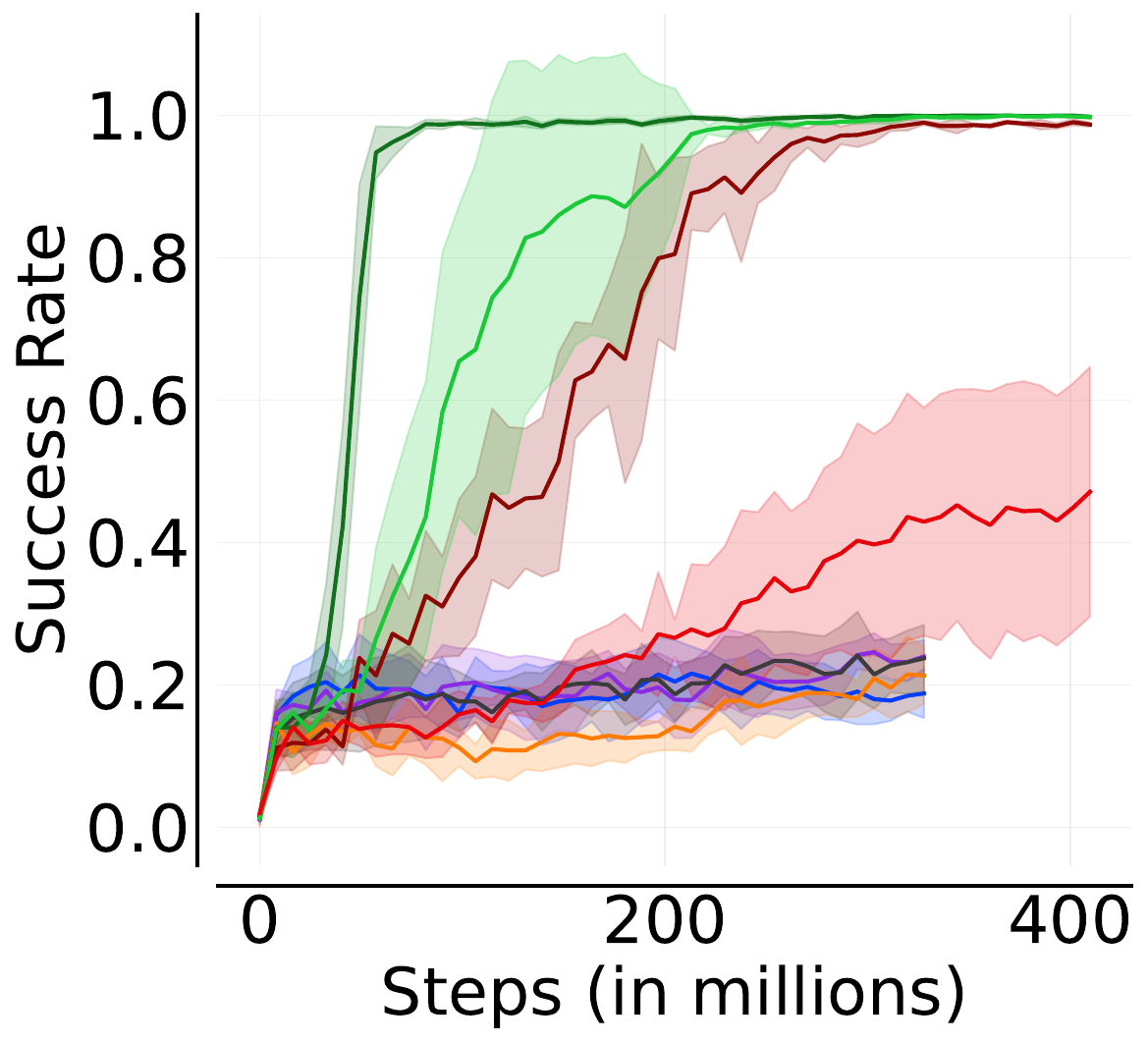}\label{fig:res_ss_2}}

  \caption{Performance comparison of several minor baselines on instances of Memory Gym's finite environments. Mortar Mayhem Act Grid (a) provides commands as a fully observable vector, bypassing the Clue Task. Both (a) and Mortar Mayhem Grid (b) utilize grid-like locomotion. Mystery Path Grid operates on grid-like locomotion, with (d) rendering the the goal and origin to the agent's observation. (e) hides these. Searing Spotlights (f) is not varied, while the minor baselines leverage observation reconstruction (Obs. Rec.) in this environment.}
  \label{fig:results_finite_2}
\end{figure}

This set of experiments reaffirms that agents trained on Memory Gym require memory.
To show this, we train several naive baselines on simplified versions of the environments.
These baselines include PPO without memory, PPO with frame stacking (4 and 16 grayscale frames), and PPO with absolute positional encoding \citep{Vaswani2017attention} provided as vector observation.
Note that the hyperparameters for these baselines are not tuned.
Figure \ref{fig:results_finite_2} shows the mean and standard deviation of 5 independent seeds for each presented agent.

In Mortar Mayhem Act Grid, where the commands are completely provided as vector observation, PPO without memory is ineffective, while the frame stacking agents achieve nearly either 4.7 or 7.3 commands with a slight upward trend, while the one with absolute positional encoding follows closely, completing 9.6 commands.
The positional encoding baseline adds the episode's current time step to the agent's observation allowing for temporal awareness within the sequential execution of commands.
Once the complete task is present (Figure \ref{fig:res_mm_grid_2}), only GRU and TrXL prove to be effective.

We obtain a similar impression when training on Mystery Path Grid where the origin and goal are not perceivable by the agent (Figure \ref{fig:res_mp_grid_2}).
If the origin and the goal are visible, a horizon of 16 frames is sufficient to train an effective policy as shown by the frame stacking agent (Figure \ref{fig:res_mp_grid_cues_2}).
Stacking 4 frames or leveraging positional encoding leads to a success rate of 42\% and 49\% respectively.
The memory-less agent's rate goes up to about 29\%.

Limited success is proven by the minor baselines in Searing Spotlights as seen in Figure \ref{fig:res_ss_2}.
Only in this experiment, the naive baselines also leverage observation reconstruction.

\section{GTrXL and LSTM Results on the Finite Environments}
\label{sec:gtrxl_lstm}

\begin{figure}[h]
    \centering
    \subfigure[Mortar Mayhem]{\includegraphics[width=0.475\textwidth]{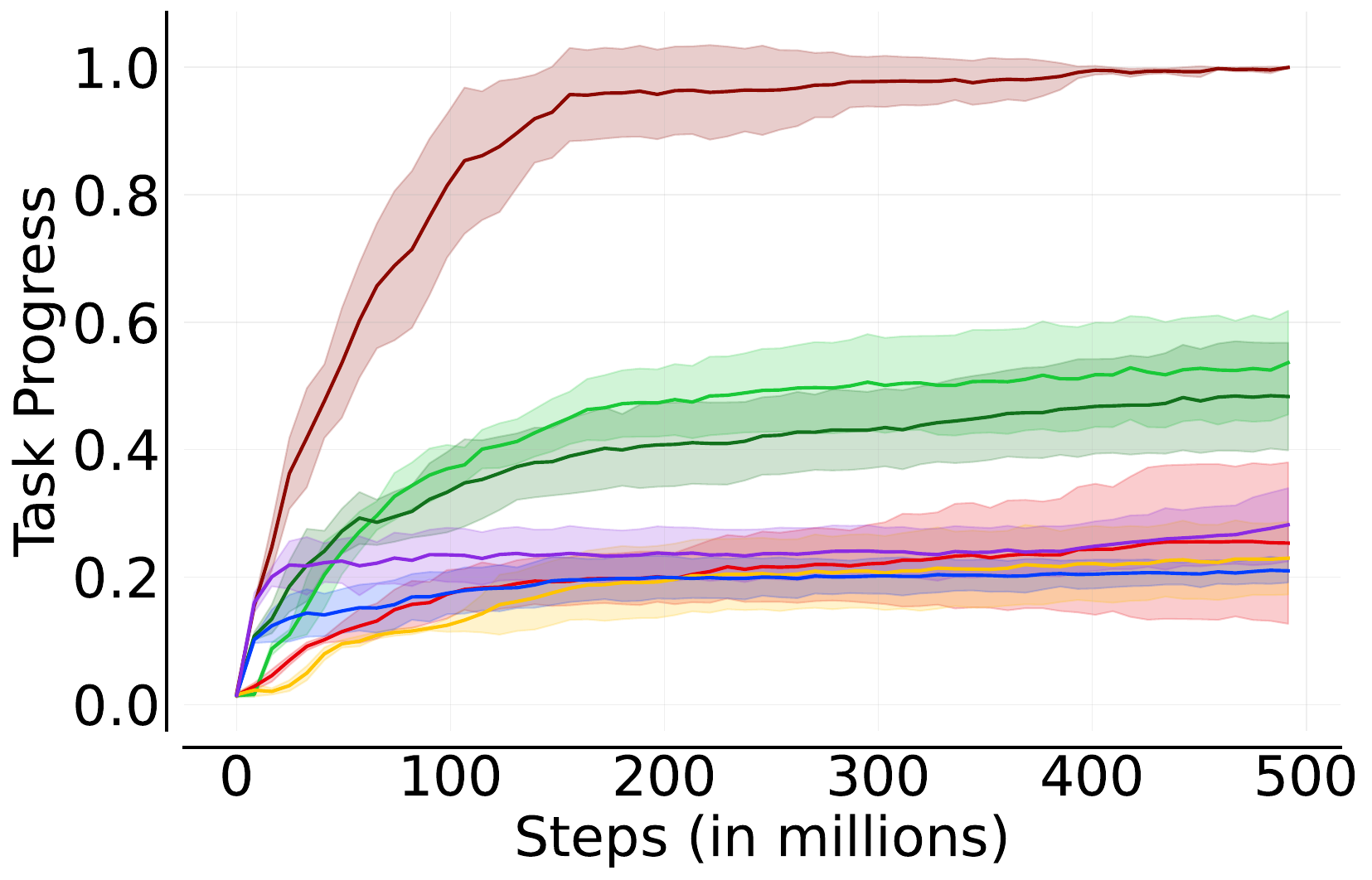}\label{fig:res_mm10_annex}}
    \hfill
    \subfigure[Mystery Path]{\includegraphics[width=0.475\textwidth]{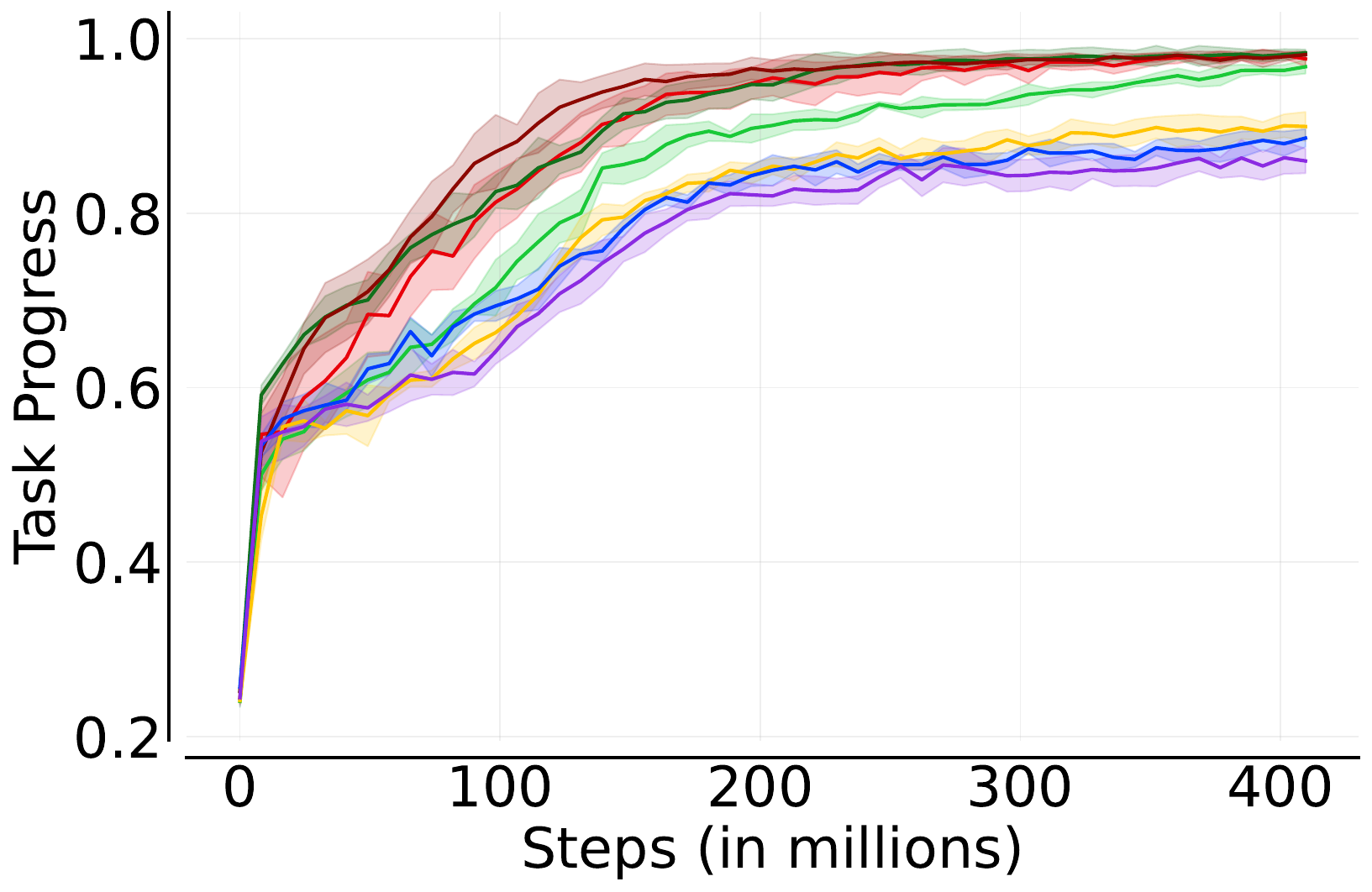}}\label{fig:res_mp_annex}

    \subfigure[Searing Spotlights]{\includegraphics[width=0.475\textwidth]{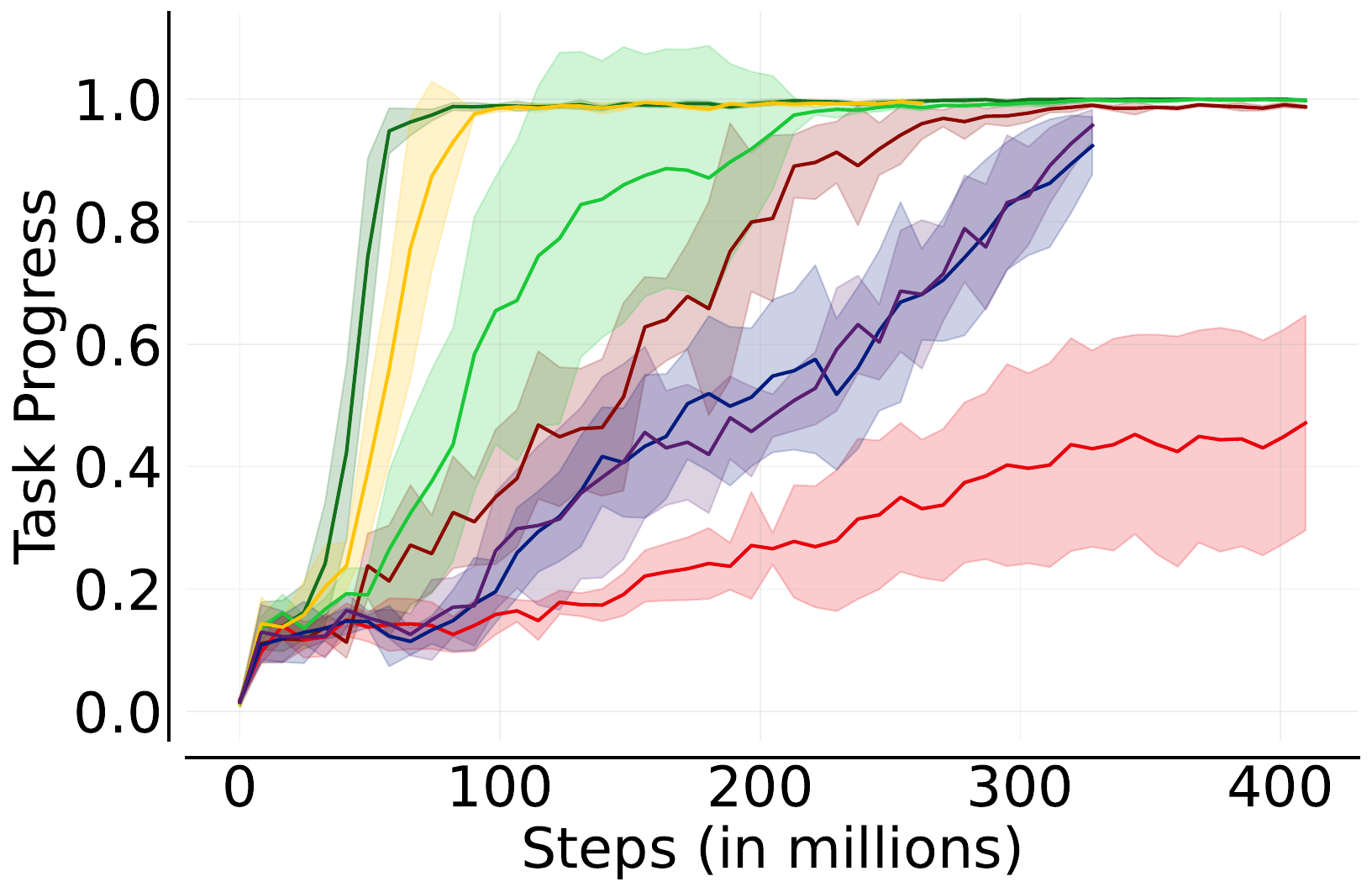}\label{fig:res_ss_annex}}
    \subfigure{\raisebox{0.4cm}{\includegraphics[width=0.26\textwidth]{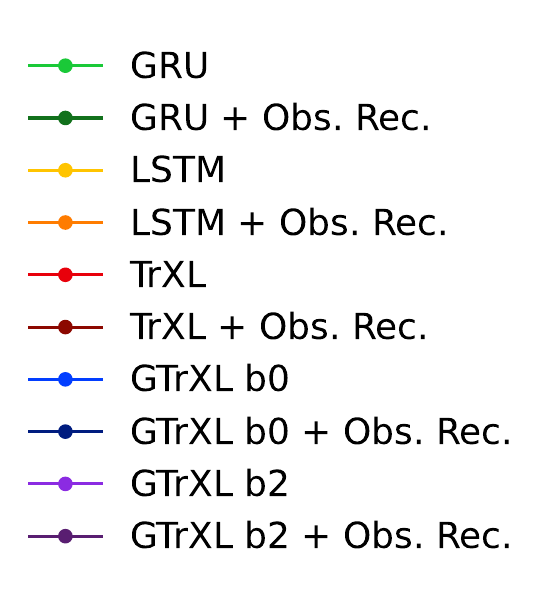}\label{fig:res_legend_annex}}}\hfill

  \caption{GTrXL and LSTM results on the finite environments.}
  \label{fig:results_finite_annex}
\end{figure}

Figure \ref{fig:results_finite_annex} presents extended results for the finite environments, showing the mean and standard deviation across 5 independent runs for each trained agent.
The performance metrics for TrXL and GRU agents remain consistent with those depicted in Figure \ref{fig:results_finite}.
We include results from agents utilizing Gated Transformer-XL (GTrXL) \citep{parisotto2020} and the Long Short-Term Memory (LSTM) \citep{Hochreiter1997LSTM}.
These models are benchmarked without any hyperparameter tuning.
For GTrXL, biases of both 0 and 2 are tested, based on the suggestion by \cite{parisotto2020} that a larger bias might enhance learning.
All additional agents in the Searing Spotlights environment employ the observation reconstruction loss (Obs. Rec.), while it is omitted in the Mystery Path environment.

For Mortar Mayhem (as seen in Figure \ref{fig:res_mm10_annex}), the LSTM agent settles at an average of 2.2 commands, whereas the GTrXL agents achieve a mean of 2.1 and 2.8 commands.
In the Mystery Path results (Figure \ref{fig:res_mp_annex}), the LSTM agent completes 90\% of the task.
The GTrXL agent with a bias of 0 reaches a 88\% progression, while the one with a bias of 2 achieves 86\%, challenging the earlier assertion by \cite{parisotto2020}.
In the Searing Spotlights environment, the LSTM agent with observation reconstruction is slightly less efficient than its GRU counterpart.
However, GTrXL agents, though still growing to success, lag behind TrXL with observation reconstruction.

We avoid making conclusive statements based on these results.
This caution stems from the fact that the LSTM and GTrXL baselines are not tuned.

\section{Wall-Time Efficiency}
\label{sec:wall_time}
    
\input{tables/walltime}

Reliably reporting the wall-time of all conducted experiments is not feasible due to varying circumstances as varying hardware or different loads of the file system.
Therefore, we only provide a selection of measurements on Endless Mortar Mayhem and Mystery Path.

Table \ref{tab:walltime} presents wall-time efficiency metrics for various agents trained on Endless Mortar Mayhem, utilizing the noctua2 high-performance cluster\footnote{\url{https://pc2.uni-paderborn.de/de/hpc-services/available-systems/noctua2}}.
We leverage 32 CPU cores (AMD Milan 7763), an NVIDIA A100 GPU, and 100GB RAM for one training run.
With PyTorch version 2, models can be lazily compiled at training onset, reducing TrXL's training time by a significant 30\%.
Previously, plain TrXL training averaged 110 hours, but this has been reduced to 76 hours.
Incorporating observation reconstruction adds roughly 3 hours, and with ground truth estimation, it extends to 81 hours.
The GRU agents are the most wall-time efficient, completing in approximately 57 hours.
Thus, recurrence emerges as the most effective and efficient architecture in both sample and wall-time metrics for the endless environments.

\begin{figure}
    \centering
    \subfigure{\includegraphics[width=0.71\textwidth]{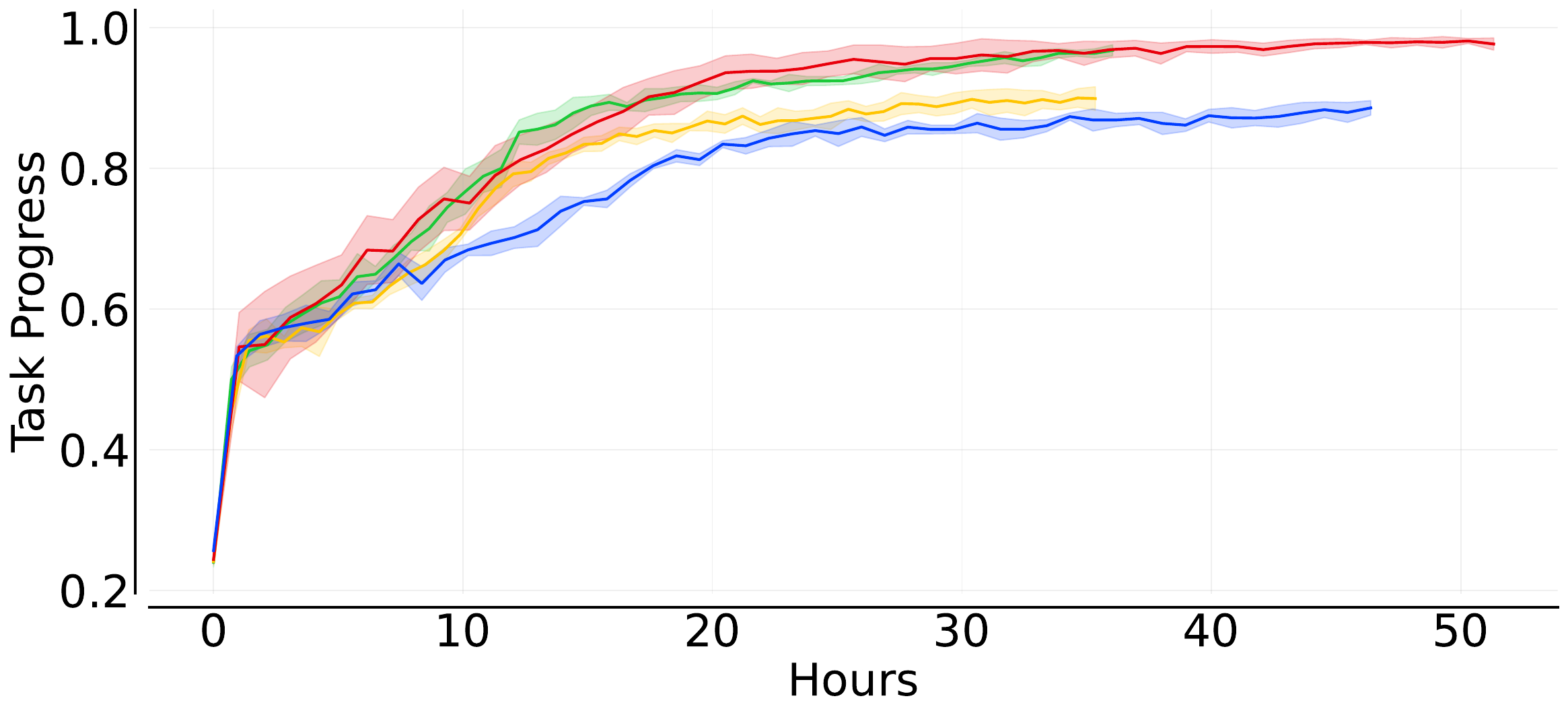}\label{fig:mp_walltime}}
    \subfigure{\raisebox{1.6cm}{\includegraphics[width=0.175\textwidth]{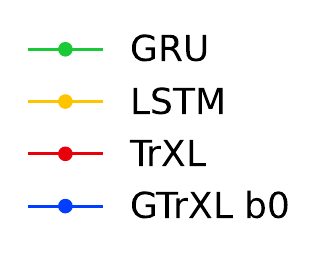}\label{fig:res_legend_annex_2}}}\hfill
    \caption{Wall-time efficiency curves retrieved from Mystery Path}
    \label{fig:walltime_mp}
\end{figure}

We further take a look at the wall-time of various agents trained on Mystery Path (Figure \ref{fig:walltime_mp}).
It becomes apparent that TrXL and GRU are quite on par concerning wall-time efficiency, whereas TrXL needs fewer samples as shown in section \ref{sec:finite_results}.
On first sight, it seems surprising that GTrXL is faster than TrXL (Table \ref{tab:walltime_mp}) even though GTrXL is more complex.
This is due to the more expensive resets of the environment.
Better policies have shorter episodes and thus reset more frequently impairing wall-time.
So as GTrXL's and LSTM's policies are inferior, their wall-time is faster.
These results were obtained from the LiDo3 high-performance cluster\footnote{\url{https://lido.itmc.tu-dortmund.de/}}.
Each run utilized 32 cores (AMD Epyc 7542), an A100 GPU, and 200GB RAM.

\input{tables/walltime_mp}

\newpage
\section{Simulation Speed of the Environments}
\label{sec:sim_speed}

Memory Gym's environments are implemented using Python and PyGame\footnote{\url{https://www.pygame.org}}.
The simulation speeds are shown in Table \ref{tab:sim_speed}.
At first glance, the endless variants appear slower than the finite ones.
This discrepancy is attributed to the shorter episodes in endless variants, which rely on less effective constant action policies, thereby increasing the impact of environment reset costs.
Compared to related memory benchmarks, our environments perform well, with only Procgen \citep{ProcgenCobbe202} being faster \citep{pleines2023memory}.
The training costs in our experiments are primarily due to the inference time of the baseline models.

\input{tables/sim_speed}






        

%% file: tables/env_parameters.tex
\begin{table}[h]
\centering
\caption{Default reset parameters of the finite environments.
Parameters marked with an asterisk (*) indicate uniform sampling.
Values enclosed in square brackets represent discrete choices, while values in parentheses denote a range from which sampling is performed.
For Mortar Mayhem Grid and Mystery Path Grid, the parameters remain the same as its parent, with only the modified ones presented.}
\small
\begin{tabular}{|lr|lr|}
\hline
\rowcolor[HTML]{EFEFEF} 
\multicolumn{2}{|c|}{\cellcolor[HTML]{EFEFEF}\textbf{Mortar Mayhem}}                  & \multicolumn{2}{c|}{\cellcolor[HTML]{EFEFEF}\textbf{Searing Spotlights}} \\
\rowcolor[HTML]{EFEFEF} 
\textbf{Parameter}                         & \textbf{Default}                         & \textbf{Parameter}                     & \textbf{Default}                \\ \hline
Agent Scale                                & 0.25                                     & Max Episode Length                     & 256                             \\
Agent Speed                                & 3                                        & Agent Scale                            & 0.25                            \\
Arena Size                                 & 5                                        & Agent Speed                            & 3                               \\
No. Available Commands                     & 9                                        & Agent Always Visible                   & False                           \\
No. Commands*                              & {[}10{]}                                 & Agent Health                           & 5                               \\
Command Show Duration*                     & {[}3{]}                                  & Sample Agent Position                  & True                            \\
Command Show Delay*                        & {[}1{]}                                  & Use Exit                               & True                            \\
Execution Duration*                        & {[}6{]}                                  & Exit Scale                             & 0.5                             \\
Execution Delay*                           & {[}18{]}                                 & Exit Visible                           & False                           \\
Show Visual Feedback                       & True                                     & Number of Coins*                       & {[}1{]}                         \\
Reward Command Failure                     & 0                                        & Coin Scale                             & 0.375                           \\
Reward Command Success                     & 0.1                                      & Coin Always Visible                    & False                           \\
Reward Episode Success                     & 0                                        & No. Initial Spotlight Spawns           & 4                               \\ \cline{1-2}
\multicolumn{2}{|c|}{\cellcolor[HTML]{EFEFEF}\textbf{Mortar Mayhem Grid}}             & No. Spotlight Spawns                   & 30                              \\
\cellcolor[HTML]{EFEFEF}\textbf{Parameter} & \cellcolor[HTML]{EFEFEF}\textbf{Default} & Spotlight Spawn Interval               & 30                              \\ \cline{1-2}
No. Available Commands                     & 5                                        & Spotlight Spawn Decay                  & 0.95                            \\
Execution Duration*                        & {[}2{]}                                  & Spotlight Spawn Threshold              & 10                              \\
Execution Delay*                           & {[}6{]}                                  & Spotlight Radius*                      & (7.5-13.75)                     \\ \cline{1-2}
\multicolumn{2}{|c|}{\cellcolor[HTML]{EFEFEF}\textbf{Mystery Path}}                   & Spotlight Speed*                       & (0.0025-0.0075)                 \\
\cellcolor[HTML]{EFEFEF}\textbf{Parameter} & \cellcolor[HTML]{EFEFEF}\textbf{Default} & Spotlight Damage                       & 1                               \\ \cline{1-2}
Max Episode Length                         & 512                                      & Light Dim Off Duration                 & 6                               \\
Agent Scale                                & 0.25                                     & Light Threshold                        & 255                             \\
Agent Speed                                & 3                                        & Show Visual Feedback                   & True                            \\
Cardinal Origin Choice*                    & {[}0, 1, 2, 3{]}                         & Show Last Action                       & True                            \\
Show Origin                                & False                                    & Show Last Positive Reward              & True                            \\
Show Goal                                  & False                                    & Render Background Black                & False                           \\
Show Visual Feedback                       & True                                     & Hide Checkered Background              & False                           \\
Reward Goal                                & 1                                        & Reward Inside Spotlight                & 0                               \\
Reward Fall Off                            & 0                                        & Reward Outside Spotlights              & 0                               \\
Reward Path Progress                       & 0.1                                      & Reward Death                           & 0                               \\
Reward Step                                & 0                                        & Reward Exit                            & 1                               \\ \cline{1-2}
\multicolumn{2}{|c|}{\cellcolor[HTML]{EFEFEF}\textbf{Mystery Path Grid}}                   & Reward Coin                            & 0.25                            \\
\cellcolor[HTML]{EFEFEF}\textbf{Parameter} & \cellcolor[HTML]{EFEFEF}\textbf{Default} & Reward Max Steps                       & 0                               \\ \cline{1-2}
Max Episode Length                         & 128                                      &                                        & \multicolumn{1}{l|}{}           \\
Reward Path Progress                       & 0                                        &                                        & \multicolumn{1}{l|}{}           \\ \hline
\end{tabular}
\normalsize
\label{tab:env_parameters}
\end{table}

%% file: tables/env_parameters_endless.tex
\begin{table}[h]
\small
\centering
\caption{Default reset parameters of the endless environments.
Parameters marked with an asterisk (*) indicate uniform sampling.
Values enclosed in square brackets represent discrete choices, while values in parentheses denote a range from which sampling is performed.
If the max episode length is equal or less than zero, episodes will not terminate because of reaching the max episode length.}
\begin{tabular}{|lr|lr|}
\hline
\rowcolor[HTML]{EFEFEF} 
\multicolumn{2}{|c|}{\cellcolor[HTML]{EFEFEF}\textbf{Endless Mortar Mayhem}}          & \multicolumn{2}{c|}{\cellcolor[HTML]{EFEFEF}\textbf{Endless Searing Spotlights}}      \\
\rowcolor[HTML]{EFEFEF} 
\textbf{Parameter}                         & \textbf{Default}                         & \cellcolor[HTML]{EFEFEF}\textbf{Parameter} & \cellcolor[HTML]{EFEFEF}\textbf{Default} \\ \hline
Max Episode Length                         & -1                                       & Max Episode Length                         & -1                                       \\
Agent Scale                                & 0.25                                     & Agent Scale                                & 0.25                                     \\
Agent Speed                                & 3                                        & Agent Speed                                & 3                                        \\
No. Available Commands                     & 9                                        & Agent Always Visible                       & False                                    \\
Command Show Duration*                     & {[}3{]}                                  & Agent Health                               & 10                                        \\
Command Show Delay*                        & {[}1{]}                                  & Sample Agent Position                      & True                                     \\
Execution Duration*                        & {[}6{]}                                  & Coin Scale                                 & 0.375                                     \\
Execution Delay*                           & {[}18{]}                                 & Coin Show Duration                         & 6                                        \\
Show Visual Feedback                       & True                                     & Coin Always Visible                        & False                                    \\
Reward Command Failure                     & 0                                        & Steps per Coin                             & 160                                      \\
Reward Command Success                     & 0.1                                      & No. Initial Spotlight Spawns               & 3                                        \\ \cline{1-2}
\multicolumn{2}{|c|}{\cellcolor[HTML]{EFEFEF}\textbf{Endless Mystery Path}}           & Spotlight Spawn Interval                   & 50                                       \\
\cellcolor[HTML]{EFEFEF}\textbf{Parameter} & \cellcolor[HTML]{EFEFEF}\textbf{Default} & Spotlight Radius*                          & (7.5-13.75)                              \\ \cline{1-2}
Max Episode Length                         & -1                                       & Spotlight Speed*                           & (0.0025-0.0075)                          \\
Agent Scale                                & 0.25                                     & Spotlight Damage                           & 1                                        \\
Agent Speed                                & 3                                        & Light Dim Off Duration                     & 6                                        \\
Show Origin                                & False                                    & Light Threshold                            & 255                                      \\
Show Past Path                             & True                                     & Show Visual Feedback                       & True                                     \\
Show Background                            & False                                    & Render Background Black                    & False                                    \\
Show Stamina                               & False                                    & Hide Checkered Background                  & False                                    \\
Show Visual Feedback                       & True                                     & Show Last Action                           & True                                     \\
Camera Offset Scale                        & 5                                        & Show Last Positive Reward                  & True                                     \\
Stamina Level                              & 20                                       & Reward Inside Spotlight                    & 0                                        \\
Reward Fall Off                            & 0                                        & Reward Outside Spotlights                  & 0                                        \\
Reward Path Progress                       & 0.1                                      & Reward Death                               & 0                                        \\
Reward Step                                & 0                                        & Reward Coin                                & 0.25                                     \\ \hline
\end{tabular}
\normalsize
\label{tab:endless_env_parameters}
\end{table}

%% file: tables/hyperparameters.tex
\begin{table}[h]
\small
\centering
\caption{Hyperparameters and architectural details used in our final experiments.
The ``Final" column denotes the selected hyperparameter values, while the ``Search Space" column represents additional discrete choices explored during the tuning process.
The last column details the old hyperparameters of our previous study \citep{pleines2023memory}.
For this column alone, empty cells indicate that the values align with those in the ``Final" column.
T-Fixup is a transformer weight initialization approach by \cite{huang2020tfixup}.
}
\begin{tabular}{|lrrrrr|}
\hline
\rowcolor[HTML]{C0C0C0} 
\multicolumn{1}{|l|}{\textbf{Hyperparameter}} & \multicolumn{1}{c|}{\textbf{Final}} & \multicolumn{3}{c|}{\textbf{Search Space}}                               & \multicolumn{1}{c|}{\textbf{Old}} \\ \hline
\multicolumn{1}{|l|}{Training Seeds}                                  & \multicolumn{1}{r|}{100000}                                 & \multicolumn{1}{r|}{}           & \multicolumn{1}{r|}{}           & \multicolumn{1}{r|}{}        &                                                           \\ \hline
\multicolumn{1}{|l|}{Number of Workers}                               & \multicolumn{1}{r|}{32}                                     & \multicolumn{1}{r|}{}           & \multicolumn{1}{r|}{}           & \multicolumn{1}{r|}{}        &                                                           \\ \hline
\multicolumn{1}{|l|}{Worker Steps}                                    & \multicolumn{1}{r|}{512}                                    & \multicolumn{1}{r|}{}           & \multicolumn{1}{r|}{}           & \multicolumn{1}{r|}{}        &                                                           \\ \hline
\multicolumn{1}{|l|}{Batch Size}                                      & \multicolumn{1}{r|}{16384}                                  & \multicolumn{1}{r|}{}           & \multicolumn{1}{r|}{}           & \multicolumn{1}{r|}{}        &                                                           \\ \hline
\multicolumn{1}{|l|}{Disocunt Factor Gamma}                           & \multicolumn{1}{r|}{0.995}                                  & \multicolumn{1}{r|}{}           & \multicolumn{1}{r|}{}           & \multicolumn{1}{r|}{}        & 0.99                                                      \\ \hline
\multicolumn{1}{|l|}{GAE Lamda}                                       & \multicolumn{1}{r|}{0.95}                                   & \multicolumn{1}{r|}{}           & \multicolumn{1}{r|}{}           & \multicolumn{1}{r|}{}        &                                                           \\ \hline
\multicolumn{1}{|l|}{Optimizer}                                       & \multicolumn{1}{r|}{AdamW}                                  & \multicolumn{1}{r|}{}           & \multicolumn{1}{r|}{}           & \multicolumn{1}{r|}{}        &                                                           \\ \hline
\multicolumn{1}{|l|}{Epochs}                                          & \multicolumn{1}{r|}{3}                                      & \multicolumn{1}{r|}{2}          & \multicolumn{1}{r|}{4}          & \multicolumn{1}{r|}{}        &                                                           \\ \hline
\multicolumn{1}{|l|}{Number of Mini Batches}                          & \multicolumn{1}{r|}{8}                                      & \multicolumn{1}{r|}{4}          & \multicolumn{1}{r|}{}           & \multicolumn{1}{r|}{}        &                                                           \\ \hline
\multicolumn{1}{|l|}{Advantage Normalization}                         & \multicolumn{1}{r|}{No}                                     & \multicolumn{1}{r|}{Batch}      & \multicolumn{1}{r|}{Mini Batch} & \multicolumn{1}{r|}{}        & Mini Batch                                                \\ \hline
\multicolumn{1}{|l|}{Clip Range Epsilon}                              & \multicolumn{1}{r|}{0.1}                                    & \multicolumn{1}{r|}{0.2}        & \multicolumn{1}{r|}{0.3}        & \multicolumn{1}{r|}{}        & 0.2                                                       \\ \hline
\multicolumn{1}{|l|}{Value Loss Coefficient}                          & \multicolumn{1}{r|}{0.5}                                    & \multicolumn{1}{r|}{0.25}       & \multicolumn{1}{r|}{}           & \multicolumn{1}{r|}{}        & 0.25                                                      \\ \hline
\multicolumn{1}{|l|}{Initial Learning Rate}                           & \multicolumn{1}{r|}{2.75e-4}                                & \multicolumn{1}{r|}{2.0e-4}     & \multicolumn{1}{r|}{3.0e-4}     & \multicolumn{1}{r|}{3.5e-4}  & 3.0e-4                                                    \\ \hline
\multicolumn{1}{|l|}{Final Learning Rate}                             & \multicolumn{1}{r|}{1.0e-5}                                 & \multicolumn{1}{r|}{}           & \multicolumn{1}{r|}{}           & \multicolumn{1}{r|}{}        & 1.0e-4                                                    \\ \hline
\multicolumn{1}{|l|}{Initial Entropy Coef.}                           & \multicolumn{1}{r|}{1.0e-4}                                 & \multicolumn{1}{r|}{1.0e-3}     & \multicolumn{1}{r|}{1.0e-2}     & \multicolumn{1}{r|}{}        &                                                           \\ \hline
\multicolumn{1}{|l|}{Final Entropy Coef.}                             & \multicolumn{1}{r|}{1.0e-6}                                 & \multicolumn{1}{r|}{}           & \multicolumn{1}{r|}{}           & \multicolumn{1}{r|}{}        & 1.0e-5                                                    \\ \hline
\multicolumn{1}{|l|}{Reconstruction Loss Coef.}                       & \multicolumn{1}{r|}{0.1}                                    & \multicolumn{1}{r|}{0.5}        & \multicolumn{1}{r|}{1.0}        & \multicolumn{1}{r|}{}        & n/a                                                       \\ \hline
\multicolumn{1}{|l|}{Maximum Gradient Norm}                           & \multicolumn{1}{r|}{0.25}                                   & \multicolumn{1}{r|}{0.35}       & \multicolumn{1}{r|}{0.5}        & \multicolumn{1}{r|}{1.0}     & 0.5                                                       \\ \hline
\rowcolor[HTML]{C0C0C0} 
\multicolumn{6}{|c|}{\textbf{Recurrent Neural Network}}                                                                                                                                                                                                                    \\ \hline
\multicolumn{1}{|l|}{Number of Recurrent Layers}                      & \multicolumn{1}{r|}{1}                                      & \multicolumn{1}{r|}{2}          & \multicolumn{1}{r|}{}           & \multicolumn{1}{r|}{}        &                                                           \\ \hline
\multicolumn{1}{|l|}{Embedding Layer}                                 & \multicolumn{1}{r|}{Yes}                                    & \multicolumn{1}{r|}{}           & \multicolumn{1}{r|}{}           & \multicolumn{1}{r|}{}        & No                                                        \\ \hline
\multicolumn{1}{|l|}{Layer Type}                                      & \multicolumn{1}{r|}{GRU}                                    & \multicolumn{1}{r|}{LSTM}       & \multicolumn{1}{r|}{}           & \multicolumn{1}{r|}{}        &                                                           \\ \hline
\multicolumn{1}{|l|}{Residual}                                        & \multicolumn{1}{r|}{False}                                  & \multicolumn{1}{r|}{True}       & \multicolumn{1}{r|}{}           & \multicolumn{1}{r|}{}        &                                                           \\ \hline
\multicolumn{1}{|l|}{Sequence Length}                                 & \multicolumn{1}{r|}{512 or max}                             & \multicolumn{1}{r|}{}           & \multicolumn{1}{r|}{}           & \multicolumn{1}{r|}{}        &                                                           \\ \hline
\multicolumn{1}{|l|}{Hidden State Size}                               & \multicolumn{1}{r|}{512}                                    & \multicolumn{1}{r|}{256}        & \multicolumn{1}{r|}{384}        & \multicolumn{1}{r|}{}        &                                                           \\ \hline
\rowcolor[HTML]{C0C0C0} 
\multicolumn{6}{|c|}{\textbf{Transformer-XL}}                                                                                                                                                                                                                              \\ \hline
\multicolumn{1}{|l|}{Number of TrXL Layers}                           & \multicolumn{1}{r|}{3}                                      & \multicolumn{1}{r|}{2}          & \multicolumn{1}{r|}{4}          & \multicolumn{1}{r|}{}        & \multicolumn{1}{l|}{}                                     \\ \hline
\multicolumn{1}{|l|}{TrXL Block Dimension}                                 & \multicolumn{1}{r|}{384}                                    & \multicolumn{1}{r|}{256}        & \multicolumn{1}{r|}{512}        & \multicolumn{1}{r|}{}        & \multicolumn{1}{l|}{}                                     \\ \hline
\multicolumn{1}{|l|}{Block Weight Initialization}                     & \multicolumn{1}{r|}{Xavier}                                 & \multicolumn{1}{r|}{Orthogonal} & \multicolumn{1}{r|}{Kaiming}    & \multicolumn{1}{r|}{T-Fixup} & \multicolumn{1}{l|}{}                                     \\ \hline
\multicolumn{1}{|l|}{Positional Encoding}                             & \multicolumn{1}{r|}{Absolute}                               & \multicolumn{1}{r|}{None}       & \multicolumn{1}{r|}{Learned}    & \multicolumn{1}{r|}{}        & \multicolumn{1}{l|}{}                                     \\ \hline
\multicolumn{1}{|l|}{Number of Attention Heads}                       & \multicolumn{1}{r|}{4}                                      & \multicolumn{1}{r|}{8}          & \multicolumn{1}{r|}{}           & \multicolumn{1}{r|}{}        & \multicolumn{1}{l|}{}                                     \\ \hline
\multicolumn{1}{|l|}{Memory Window Length}                            & \multicolumn{1}{r|}{256 or max}                             & \multicolumn{1}{r|}{}           & \multicolumn{1}{r|}{}           & \multicolumn{1}{r|}{}        & \multicolumn{1}{l|}{}                                     \\ \hline
\end{tabular}
\label{tab:hyperparameters}
\end{table}

%% file: tables/walltime.tex
\begin{table}[h]
\centering
\caption{Wall-time efficiency metrics for various agents trained on Endless Mortar Mayhem. Agents are arranged from left to right based on increasing mean training duration, measured in hours. Other values are represented in seconds and comprise the time needed for a single update (i.e. one complete iteration of running PPO). The second row indicates if the neural network was compiled before training. Note: TrXL* metrics use 4k initial training data points, while other runs utilize 150k data points covering the full training process.}
\begin{tabular}{l|
>{}r r
>{}r r
>{}r r|}
\cline{2-7}
                                    & \multicolumn{1}{l}{GRU} & \multicolumn{1}{l}{TrXL} & \multicolumn{1}{l}{\begin{tabular}[c]{@{}l@{}}TrXL +\\ QPos\end{tabular}} & \multicolumn{1}{l}{\begin{tabular}[c]{@{}l@{}}TrXL +\\ Obs. Rec.\end{tabular}} & \multicolumn{1}{l}{\begin{tabular}[c]{@{}l@{}}TrXL +\\ LR + QPos + GT\end{tabular}} & \multicolumn{1}{l|}{TrXL*} \\ \hline
\multicolumn{1}{|l|}{Compiled}      & No                                              & Yes                      & Yes                                                                                               & Yes                                                                            & Yes                                                                                                         & No                        \\ \hline
\multicolumn{1}{|l|}{Mean}          & 4.11                                            & 5.53                     & 5.6                                                                                               & 5.76                                                                           & 5.86                                                                                                        & 7.97                      \\
\multicolumn{1}{|l|}{Std}           & 0.35                                            & 1.12                     & 1.31                                                                                              & 1.1                                                                            & 1.68                                                                                                        & 0.25                      \\
\multicolumn{1}{|l|}{Min}           & 3.6                                             & 5.2                      & 5                                                                                                 & 5.4                                                                            & 5.3                                                                                                         & 7.6                       \\
\multicolumn{1}{|l|}{Median}        & 4.1                                             & 5.4                      & 5.5                                                                                               & 5.7                                                                            & 5.8                                                                                                         & 7.9                       \\
\multicolumn{1}{|l|}{IQM}           & 4.03                                            & 5.48                     & 5.53                                                                                              & 5.72                                                                           & 5.82                                                                                                        & 7.93                      \\ \hline
\multicolumn{1}{|l|}{Mean Duration} & 57.08                                           & 76.81                    & 77.78                                                                                             & 80                                                                             & 81.39                                                                                                       & 110.69                    \\ \hline
\end{tabular}
\label{tab:walltime}
\end{table}

%% file: tables/walltime_mp.tex
\begin{table}[h]
\centering
\caption{Wall-time efficiency metrics for various agents trained on Mytsery Path. Agents are arranged from left to right based on increasing mean training duration, measured in hours. Other values are represented in seconds and comprise the time needed for a single update (i.e. one complete iteration of running PPO). The second row indicates if the neural network was compiled before training.}
\begin{tabular}{l
>{}r r
>{}r r|}
\cline{2-5}
\multicolumn{1}{l|}{}              & \multicolumn{1}{c}{LSTM} & \multicolumn{1}{c}{GRU} & \multicolumn{1}{c}{GTrXL B0} & \multicolumn{1}{c|}{TrXL} \\ \hline
\multicolumn{1}{|l|}{Compiled}      & No                                               & No                      & Yes                                                  & Yes                       \\ \hline
\multicolumn{1}{|l|}{Mean}          & 5.09                                             & 5.19                    & 6.68                                                 & 7.39                      \\
\multicolumn{1}{|l|}{Std}           & 0.22                                             & 0.28                    & 0.31                                                 & 0.24                      \\
\multicolumn{1}{|l|}{Min}           & 4.2                                              & 4.0                     & 5.8                                                  & 6.4                       \\
\multicolumn{1}{|l|}{Max}           & 7.6                                              & 7.0                     & 62.2                                                 & 11.5                      \\
\multicolumn{1}{|l|}{Median}        & 5.1                                              & 5.2                     & 6.6                                                  & 7.4                       \\
\multicolumn{1}{|l|}{IQM}           & 5.09                                             & 5.18                    & 6.64                                                 & 7.39                      \\ \hline
\multicolumn{1}{|l|}{Mean Duration} & 35.35                                            & 36.04                   & 46.4                                                 & 51.32                     \\ \hline
\end{tabular}
\label{tab:walltime_mp}
\end{table}

%% file: tables/sim_speed.tex
\begin{table}[h!]
\centering
\caption{Comparison of the simulation speed of Memory Gym's environments. A constant action is executed to measure the speed, and the episode reset duration is included in the steps per second. The measurements were done on an AMD Ryzen 7 2700X, aggregating 1000 episodes per environment.}
\begin{tabular}{lrr}
\hline
\textbf{Environment}                & \textbf{Mean (steps/sec)} & \textbf{Std (steps/sec)} \\ \hline
Mystery Path                        & 14033 & 158 \\ \hline
Mortar Mayhem                       & 13187 & 208 \\ \hline
Endless Mortar Mayhem               & 9330  & 452 \\ \hline
Endless Mystery Path                & 7278  & 555 \\ \hline
Endless Searing Spotlights          & 5828  & 106 \\ \hline
Searing Spotlights                  & 5789  & 165 \\ \hline
\end{tabular}
\label{tab:sim_speed}
\end{table}